%% file: main.tex
\documentclass[]{academic_template}

\usepackage[toc,page,header]{appendix}

\usepackage[utf8]{inputenc}

\usepackage{amsmath}
\usepackage{amssymb}
\usepackage{amsfonts}

\usepackage{graphicx}
\usepackage{booktabs}
\usepackage{makecell}
\usepackage{multirow}
\usepackage{tabularx}
\usepackage{array}
\usepackage{wrapfig}
\usepackage{caption}
\usepackage{subcaption}

\usepackage[table,dvipsnames,xcdraw]{xcolor}

\usepackage{algorithm}
\usepackage{algpseudocode}

\usepackage{listings}
\usepackage{pifont}
\usepackage[most]{tcolorbox}
\usepackage{enumitem}
\usepackage{nicefrac}
\usepackage{microtype}

\usepackage{hyperref}

\newcommand{\ourmethod}{\textsc{Moment-Video}}

\definecolor{darkgreen}{RGB}{0,120,0}
\definecolor{darkred}{RGB}{180,0,0}
\definecolor{airforcered}{rgb}{0.7, 0.2, 0.2}

\newcommand{\cmark}{\textcolor{darkgreen}{\ding{51}}}
\newcommand{\xmark}{\textcolor{darkred}{\ding{55}}}

\setlogoheight{14mm}   %
\setlogospacing{5mm}
\setsjtublue %

\settitlerulethickness{3pt}  %

\abstractboxon %
\setabstractframecolor{gray} %
\setabstractbgcolor{gray!10} %
\setlogotolineshift{5mm} %

\settoprulethickness{2.5pt}  
\setbottomrulethickness{1.5pt} 

\title{Moment-Video: Diagnosing Temporal Fidelity of Video MLLMs on Momentary Visual Events}

\author[2,3,*]{Xiaolin Liu}
\author[4,*]{Yilun Zhu}
\author[1,*,\ddagger]{Xiangyu Zhao}
\author[1]{Xuehui Wang}
\author[1]{Yan Li}
\author[5]{Xin Li}
\author[6]{Kai Wang}
\author[5]{Haoyu Cao}
\author[2]{Shaofeng Zhang}
\author[4]{Xu Yang}
\author[1]{Zhihang Zhong}
\author[1, \dagger]{Xue Yang}

\affiliation[1]{Shanghai Jiao Tong University}
\affiliation[2]{University of Science and Technology of China}
\affiliation[3]{Beijing Zhongguancun Academy}
\affiliation[4]{Southeast University}
\affiliation[5]{Tencent}
\affiliation[6]{National University of Singapore}

\contribution[*]{Equal contribution}
\contribution[\ddagger]{Project Lead}
\contribution[\dagger]{Corresponding Author}

\abstract{
Video multimodal large language models (MLLMs) have made rapid progress in general and long-form video understanding, yet their ability to capture and use brief, answer-critical visual evidence remains underexplored. Momentary actions or state transitions may determine an answer while occupying only a small fraction of the video. Performance on broad video understanding tasks alone does not establish whether models can reliably identify and use this temporally localized evidence. We introduce \ourmethod{}, a benchmark for evaluating temporal fidelity through momentary visual event understanding. Questions are constructed around visually observable events or event sequences, requiring models to notice, count, describe, or reason about transient evidence. The benchmark contains 1,000 human-verified video-QA pairs across seven domains and 25 subcategories, covering four tasks: Temporal Occurrence, Temporal Counting, Action Description, and Temporal Reasoning. We evaluate 34 proprietary and open-source MLLMs. Under the main evaluation settings, the best-performing model, Seed-2.0-Pro, achieves 39.6\% overall accuracy, while most open-source models remain below 25\%. Diagnostic analyses show that denser frame sampling improves some models, but substantial errors persist in an audited subset even when human review judges the sampled visual evidence sufficient to answer. Performance also declines on longer videos, suggesting additional temporal-localization challenges. By exposing these limitations, \ourmethod{} provides a diagnostic benchmark to guide the evaluation and development of video MLLMs that more reliably understand brief, answer-critical visual events.
}

\checkdata[Project Page]{\url{https://moment-video.netlify.app/}}
\checkdata[Code]{\url{https://github.com/VisionXLab/Moment-Video}}
\checkdata[Hugging Face]{\url{https://huggingface.co/datasets/VisionXLab/Moment-Video}}

\begin{document}
\maketitle

\input{sections/introduction_v2}

\input{sections/related_work}

\input{sections/benchmark_v2}

\input{sections/experiments}

\input{sections/conclusion}

\subsection*{AI use statement}

We used generative AI tools to polish the manuscript for grammar,
clarity, and readability. All AI-assisted edits were reviewed and
verified by the authors for technical accuracy and consistency
with the underlying research. As described in Appendix~\ref{appd:data_quality_control}, we also used
Gemini-3-Flash-Preview to screen manually annotated samples for
potentially unclear questions, missing referents, ambiguous answers,
inconsistent task labels, and possible shortcuts. The model only
flagged potential issues; human annotators verified the samples and
made the final decisions to accept, revise, or remove them.
The screening model did not generate ground-truth answers or
determine their final correctness. For evaluation, we used an LLM to assess the semantic equivalence
between open-ended model predictions and human-verified reference
answers, following the protocol in
Section~\ref{sec:answer_format_scoring} and the prompt in
Appendix~\ref{appendix_prompt}. The evaluator did not receive the
videos and was instructed to base its judgments only on the supplied
reference answer and model prediction, without introducing
external knowledge. We assessed its agreement with human judgments
as reported in Appendix~\ref{appendix_judgeValid}.

The authors reviewed all AI-assisted contributions for technical
accuracy, appropriate attribution, and consistency with the underlying
data and experiments. We take responsibility for the final content
of this work, including its text, annotations, results, claims,
and associated artifacts.

\subsection*{Ethics statement}

MOMENT-VIDEO is a benchmark for evaluating video understanding and does not
involve the development or deployment of systems for high-stakes
decision-making. The benchmark is constructed from existing public datasets,
publicly accessible videos, and self-collected recordings. We followed the
applicable terms of use and licenses when collecting and processing the data.
The annotations focus exclusively on visually observable events and do not
involve the inference of sensitive personal attributes. We manually reviewed
the collected samples and excluded content containing sensitive personal
information or other inappropriate material. Any future release of the
benchmark will respect the licenses and redistribution conditions of the
original video sources.

\subsection*{Reproducibility statement}

We provide detailed documentation to support the reproducibility of
MOMENT-VIDEO and its evaluation results. Section~3 describes the benchmark
construction procedure, event-selection criteria, task taxonomy, and
evaluation protocol. The appendices provide the data sources and processing
details, dataset statistics, annotation and quality-control procedures, human
solvability results, model-specific inference settings, frame-sampling
configurations, additional experimental results, and the complete
LLM-as-a-Judge prompt. The reported model evaluations use fixed benchmark
annotations and explicitly documented input settings.

\bibliographystyle{plainnat}

\clearpage
\beginappendix
\input{sections/appendix}

\end{document}

%% file: sections/introduction_v2.tex
\section{Introduction}
\label{sec:introduction}

Video multimodal large language models (MLLMs) have recently made substantial progress in dynamic visual understanding. 
By extending image-language models to videos through frame sampling, temporal positional encoding, visual-token compression, and multimodal instruction tuning, recent systems achieve strong performance in video question answering, action recognition, temporal reasoning, and long-video comprehension~\cite{qwen3_vl,videollama3,internvl35,llava_octopus}. 
More recent omni-modal and reasoning-oriented models further enhance video understanding with audio-visual inputs, chain-of-thought reasoning, reinforcement learning, and tool-augmented inference~\cite{kimi_vl,qwen35_omni,glm45v,video_r1,videochat_r1,videochat_r15}. 
Despite these advances, the ability to use brief but decisive visual evidence remains insufficiently evaluated.

Many real-world video questions depend on \emph{momentary visual events}: visually observable actions or state transitions over a short duration, sometimes only a few frames. 
Examples include a pedestrian suddenly entering the road~\citep{event1}, a machine exhibiting momentary abnormal motion~\citep{event2}, a GUI element flashing after a command, or a game character performing a short attack that determines the outcome. 
Although these events occupy only a brief portion of the video, they are often decisive for the answer. 
Sparse sampling can omit critical frames, leaving models without direct observations of the queried event. Correct answers may then depend on indirect contextual cues. 
This motivates \emph{temporal fidelity}: the ability to capture, retain, and reason over short-lived, answer-critical visual evidence. We assess this capability through model answers, without directly probing internal representations.

\begin{figure}[t]
    \centering
    \includegraphics[width=\linewidth]{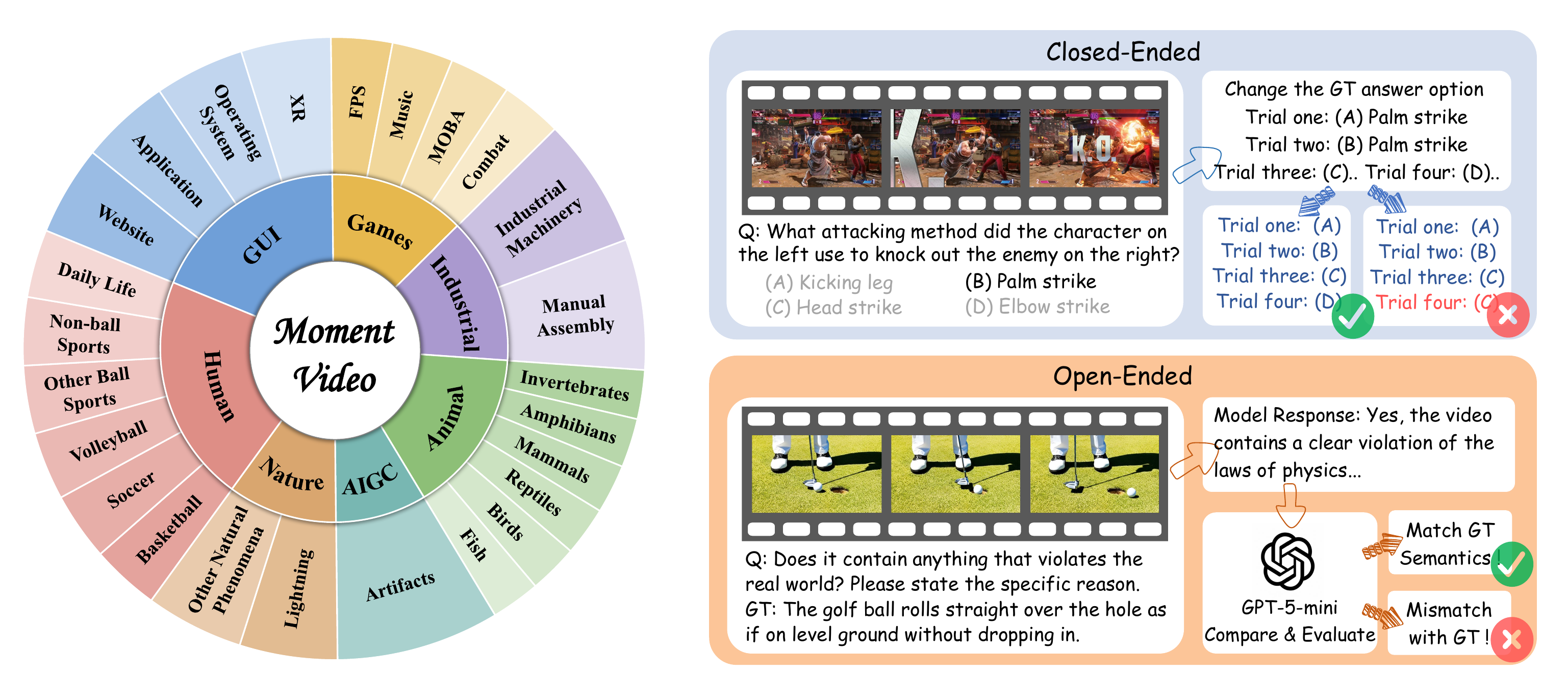}
    \caption{Overview of \textsc{Moment-Video}. Left: domain and subcategory taxonomy. Right: evaluation protocols for closed-ended and open-ended questions.}
    \label{fig:fig1}
\end{figure}

Existing video benchmarks rarely isolate this capability. 
General-purpose, long-video, and reasoning-oriented benchmarks~\cite{mvbench,video_mme,video_mme_v2,lvbench,longvideobench,mlvu,hourvideo,egoschema,all_angles_bench,crossvid,videoreasonbench,videommmu,mmvu} assess scene understanding, temporal aggregation, or persistent objects. 
Motion-centric benchmarks~\cite{motionbench,favorbench} evaluate motion perception or description, but their questions generally do not focus on brief, sampling-sensitive events. 
Thus, it remains unclear whether current MLLMs can reliably detect, count, describe, and reason about transient, momentary visual events.

To address this gap, we introduce \textsc{Moment-Video}, a benchmark for diagnosing temporal fidelity via momentary visual event understanding. 
It contains 1,000 human-verified video-QA pairs across 7 domains and 25 fine-grained subcategories, covering real-world and virtual scenarios including AI-generated content, GUI recordings, natural scenes, industrial environments, games, human activities, and animals. 
Each question is grounded in a localized momentary event, requiring answers to depend on event evidence rather than global context or external knowledge.
A small subset of questions involve transient interface text or observable animal behavior; these cues are included only when they are visually grounded and answer-critical.

Questions in \textsc{Moment-Video} are organized into four complementary task types. 
\emph{Temporal Occurrence (TO)} evaluates whether a model can notice that a brief event or state transition happens. \emph{Temporal Counting (TC)} evaluates whether a model can track and count transient actions, object changes, or repeated event occurrences. \emph{Action Description (AD)} evaluates whether a model can characterize the dynamics of a momentary event, such as its direction, trajectory, target, interaction, or state change. \emph{Temporal Reasoning (TR)} evaluates whether a model can integrate the pre-event state, the momentary event, and the post-event state to infer the event's consequence. Together, these task types form a progressive evaluation of temporal fidelity: from capturing the existence of a brief event, to preserving its dynamic details, to using it for state-change reasoning.

We evaluate 34 proprietary and open-source MLLMs. 
The best-performing model, Seed-2.0-Pro, achieves 39.6\% overall accuracy, while most open-source models remain below 25\%. 
Task-wise, AD achieves the strongest average performance, while TC is consistently the most challenging; performance on TO and TR is more model-dependent and comparable overall.
Category-wise results vary across domains; scene complexity, small objects, and fast interactions are possible contributors to these differences.

Diagnostic analyses of frame rate and video duration show that denser sampling improves some models but does not fully resolve the bottleneck. 
Accuracy is also lower on longer videos, consistent with greater difficulty locating brief evidence. 
These findings reveal a temporal fidelity limitation in current video MLLMs, motivating more fine-grained, event-aware temporal processing.

Our contributions are summarized as follows:

\noindent\textbf{(1)} We introduce \textsc{Moment-Video} to evaluate temporal fidelity through brief, answer-critical events, with 1,000 human-verified video-QA pairs across 7 domains and 25 subcategories.

\noindent\textbf{(2)} We design four complementary tasks (TO, TC, AD, TR) to assess event detection, counting, description, and reasoning over event-induced state changes.

\noindent\textbf{(3)} We evaluate 34 MLLMs and analyze frame rate and video duration. Denser sampling helps some models, yet errors persist even when human review judges the sampled evidence sufficient in an audited subset. Accuracy also declines on longer videos, suggesting additional challenges in locating brief, answer-critical evidence.

%% file: sections/related_work.tex
\section{Related Work}
\noindent\textbf{Advancements in Video MLLMs.} 
Recent video MLLMs have evolved from simple frame-based extensions of image-language models to more capable systems with temporal encoding, dynamic visual processing, visual-token compression, and large-scale multimodal instruction tuning~\citep{qwen3_vl,videollama3,internvl35,llava_octopus}. These models typically transform sampled frames or video clips into temporally ordered visual tokens and align them with LLMs for video question answering, action recognition, temporal reasoning, and long-video comprehension. Meanwhile, omni-modal and reasoning-oriented models further extend video understanding with audio-visual inputs, thinking-style reasoning, reinforcement learning, and tool-augmented inference~\citep{qwen35_omni,kimi_vl,glm45v,video_r1,videochat_r1,videochat_r15}. Despite these advances, reliably capturing and using brief, answer-critical visual events remains insufficiently evaluated, motivating the focus of \ourmethod{} on momentary-event understanding.

\noindent\textbf{Benchmarks for Video MLLMs.} 
A growing number of benchmarks evaluate video MLLMs. MVBench~\citep{mvbench}, Video-MME~\citep{video_mme}, and Video-MME-v2~\citep{video_mme_v2} assess broad video understanding across scenarios and tasks. Other benchmarks emphasize long-video comprehension~\citep{lvbench,longvideobench,mlvu,hourvideo}, egocentric understanding~\citep{egoschema}, multi-view or cross-video reasoning~\citep{all_angles_bench,crossvid}, or complex video reasoning~\citep{videoreasonbench,videommmu,mmvu}. MotionBench~\citep{motionbench} evaluates fine-grained motion perception, FAVOR-Bench~\citep{favorbench} includes closed-ended and open-ended motion understanding, and VideoAds~\citep{videoads} studies fast-paced advertisements. These settings can contain brief events; \ourmethod{} makes dependence on such evidence an explicit construction criterion. Annotators identify the target event or event sequence before writing questions and reject questions answerable without the critical moment. Across natural and virtual domains, four tasks test event detection, counting, description, and consequence inference. The distinction lies in requiring localized evidence to determine the answer, beyond selecting videos with motion or rapid scene changes. Table~\ref{tab:benchmark_comparison} summarizes the design differences.

%% file: sections/benchmark_v2.tex
\section{Benchmark Design}
\label{sec:benchmark_design}

\ourmethod{} is designed to evaluate the temporal fidelity of video MLLMs through momentary visual event understanding. The final benchmark contains 1,000 videos paired with 1,000 human-verified question-answer pairs, spanning 7 high-level domains and 25 fine-grained subcategories. We illustrate the domain taxonomy and the evaluation formats used in the benchmark in Fig.~\ref{fig:fig1}.

\subsection{What Makes an Event Momentary?}
\label{sec:momentary_definition}

We define a \emph{momentary visual event} as a brief, visually observable action or state transition whose temporally localized evidence is critical to answering the question. Missing this moment can change the answer even when the surrounding scene is understood. Here, momentariness describes temporal brevity and answer criticality, without imposing a fixed duration threshold. We randomly selected 300 samples through stratified sampling and manually annotated the start and end times of all answer-critical events, yielding 609 intervals. Of these, 87.7\% lasted at most 1 second, 97.9\% at most 2 seconds, and 71.4\% occupied at most 10\% of the corresponding video duration (Appendix~\ref{appendix_dataSta}).
To operationalize this definition during dataset construction, we impose the following criteria:

\begin{table*}[t]
\centering
\small
\setlength{\tabcolsep}{4pt}
\renewcommand{\arraystretch}{1.12}
\caption{
Comparison with existing video benchmarks along dimensions for momentary visual event understanding.
``Partial'' means the property appears in part but is not the central design target. ``Fully Human-curated’’ means video selection, question writing, and answer annotation are manual.
}
\label{tab:benchmark_comparison}
\resizebox{\textwidth}{!}{
\begin{tabular}{lcccccc}
\toprule
\textbf{Benchmark} 
& \textbf{\#Videos}
& \textbf{Primary Focus}
& \textbf{Momentary-event Focus}
& \textbf{Fully Human-curated}
& \textbf{Open-ended}
& \textbf{Frame-sampling Analysis} \\
\midrule
MVBench~\citep{mvbench}
& 4,000
& General understanding
& \xmark
& \xmark
& \xmark
& \xmark \\

Video-MME~\citep{video_mme}
& 900
& Comprehensive understanding
& \xmark
& \cmark
& \xmark
& \xmark \\

Video-MME-v2~\citep{video_mme_v2}
& 800
& Robust video understanding
& \xmark
& \cmark
& \xmark
& \xmark \\

LVBench~\citep{lvbench}
& 103
& Long-video understanding
& \xmark
& Partial
& \xmark
& \xmark \\

MotionBench~\citep{motionbench}
& 5,385
& Motion perception
& Partial
& Partial
& \xmark
& \cmark \\

FAVOR-Bench~\citep{favorbench}
& 1,776
& Motion description
& Partial
& Partial
& \cmark
& \xmark \\

VideoAds~\citep{videoads}
& 200
& Fast-paced advertisements
& Partial
& Partial
& \xmark
& \cmark \\

\midrule
\textbf{\ourmethod}
& \textbf{1,000}
& \textbf{Momentary visual events}
& \cmark
& \cmark
& \cmark
& \cmark \\
\bottomrule
\end{tabular}
}
\end{table*}

\textit{1. Localized evidence.} The decisive evidence for answering the question must be concentrated around a specific moment in the video, rather than being distributed across the entire clip. For example, the answer may depend on the instant when a ball changes direction, an object falls, a GUI element appears, or an animal suddenly moves.

\textit{2. Transient visual change.} The target event must involve a temporary action, interaction, or state transition. Questions that can be answered from a persistent object attribute, a static scene layout, or the overall topic of the video are excluded. This requirement ensures that the benchmark evaluates dynamic visual evidence rather than static image understanding.

\textit{3. Sampling sensitivity.} The event should have a non-trivial chance of being missed under sparse frame sampling.  When input frames are sampled sparsely, a brief event may fall between sampled timestamps and be absent from the visual input.

\textit{4. Evaluability.} Each question must have a well-defined answer. We use open-ended questions when the reference answer is semantically convergent and can be judged reliably. When the answer is difficult to standardize in free-form language, we use a multiple-choice format with manually designed distractors that are visually plausible and semantically close to the correct answer.

Together, these criteria distinguish \ourmethod{} from benchmarks that evaluate generic motion recognition or general video-level semantics. 
The key construction constraint is not simply that the video contains motion, but that the answer depends on brief, localized, visually grounded evidence.

\subsection{Dataset Construction}
\label{sec:dataset_construction}

We construct \ourmethod{} through a multi-stage pipeline consisting of domain taxonomy design, video collection, event-centric temporal grounding and human-expert annotation (Appendix Fig.~\ref{fig:dataset_pipeline}).

\begin{wrapfigure}{R}{0.45\linewidth}
    \centering
    \includegraphics[width=\linewidth]{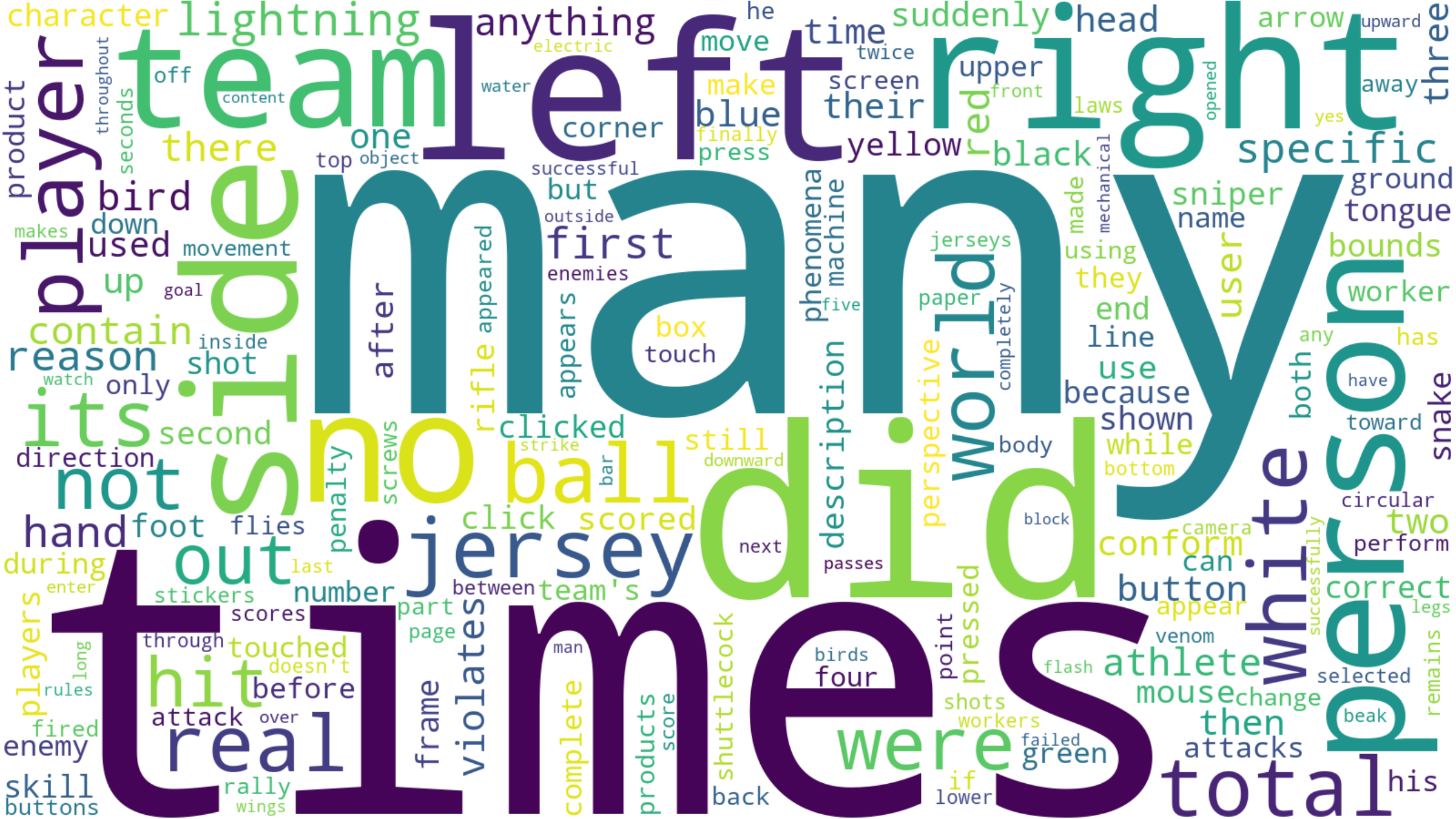}
    \caption{Keyword cloud of frequent semantic concepts in \ourmethod.}
    \label{fig:keyword_cloud}
    \vspace{-10pt}
\end{wrapfigure}
\textbf{Domain taxonomy.}
To ensure broad coverage of momentary visual events, we build a hierarchical taxonomy with 7 high-level domains: AIGC, GUI, Nature, Industry, Games, Human, and Animals. These domains cover diverse visual styles, motion patterns, scene layouts, and event types. Each domain is divided into fine-grained subcategories. For example, the Games domain includes FPS games, MOBA games, combat games, and music games; the Human domain includes ball sports, non-ball sports, and daily-life activities; and the Animals domain includes birds, mammals, reptiles, amphibians, fish, and invertebrates. This taxonomy allows us to evaluate whether models can generalize momentary event understanding across natural videos, screen recordings, synthetic videos, and complex interactive environments. The category distribution is shown in Fig.~\ref{fig:fig1}, and the keyword cloud in Fig.~\ref{fig:keyword_cloud} further illustrates the semantic coverage of \ourmethod.

\textbf{Video collection.}
We source candidate videos from public online sources~\citep{risebench,gui_xr,light,workerdata,multisports,SVW,badminton,UCF101,ani_kingdom} and self-collected recordings, each required to feature at least one visually clear momentary event. We filter out low-quality videos (e.g., severe blur, low resolution, heavy occlusion) and those relying on audio, subtitles, or external context for comprehension.
Furthermore, we revise or remove samples with persistent textual shortcuts; cropped or masked samples must pass the same screening criteria again.
The final dataset encompasses natural and synthetic domains, including real-world footage, screen recordings, gameplay, and AI-generated videos. Further details are provided in Appendix \ref{appendix_dsource}.

\textbf{Event-centric temporal grounding.}
For each candidate video, annotators first identify the target momentary event before writing the question. The same screening standards apply to all 1,000 videos, including the 842 from existing datasets and other public sources. Because many target events are brief and lack reliable frame-level boundaries, we use weak temporal grounding instead of requiring dense start and end timestamps. Annotators record the approximate critical moment, involved objects or agents, the event description, the event count when applicable, and relevant pre-event and post-event states. This information is used for annotation and quality control, but is not provided to evaluated models during testing.
When a video contains multiple brief events, annotators specify the queried event or event sequence and the evidence needed to answer the question, including multiple occurrences for counting or reasoning. A sample is rejected or revised if the event cannot be localized, if competing events yield different answers, or if the question is answerable without the key moment.

\textbf{Human-expert annotation.}
Based on localized target events, annotators formulate a question and reference answer for each video. Each sample is labeled with a domain, subcategory, task type, and answer format. Two annotators cross-check that the visual evidence determines a unique answer and that surrounding states or persistent clues alone cannot reveal it. For open-ended questions, the reference answer is a concise semantic description rather than an exhaustive video caption. For choice questions, annotators create visually plausible distractors close to the correct option, avoiding those removable by common sense, option length, or language priors. A human solvability study provides task- and domain-level reference accuracies on sampled questions (Appendix~\ref{appendix_human_valid}). Dataset statistics and annotation quality control are detailed in Appendices~\ref{appendix_dataSta} and~\ref{appd:data_quality_control}, respectively.

\subsection{Task Taxonomy}
\label{sec:task_taxonomy}

\begin{figure}[t]
    \centering
    \includegraphics[width=0.99\linewidth]{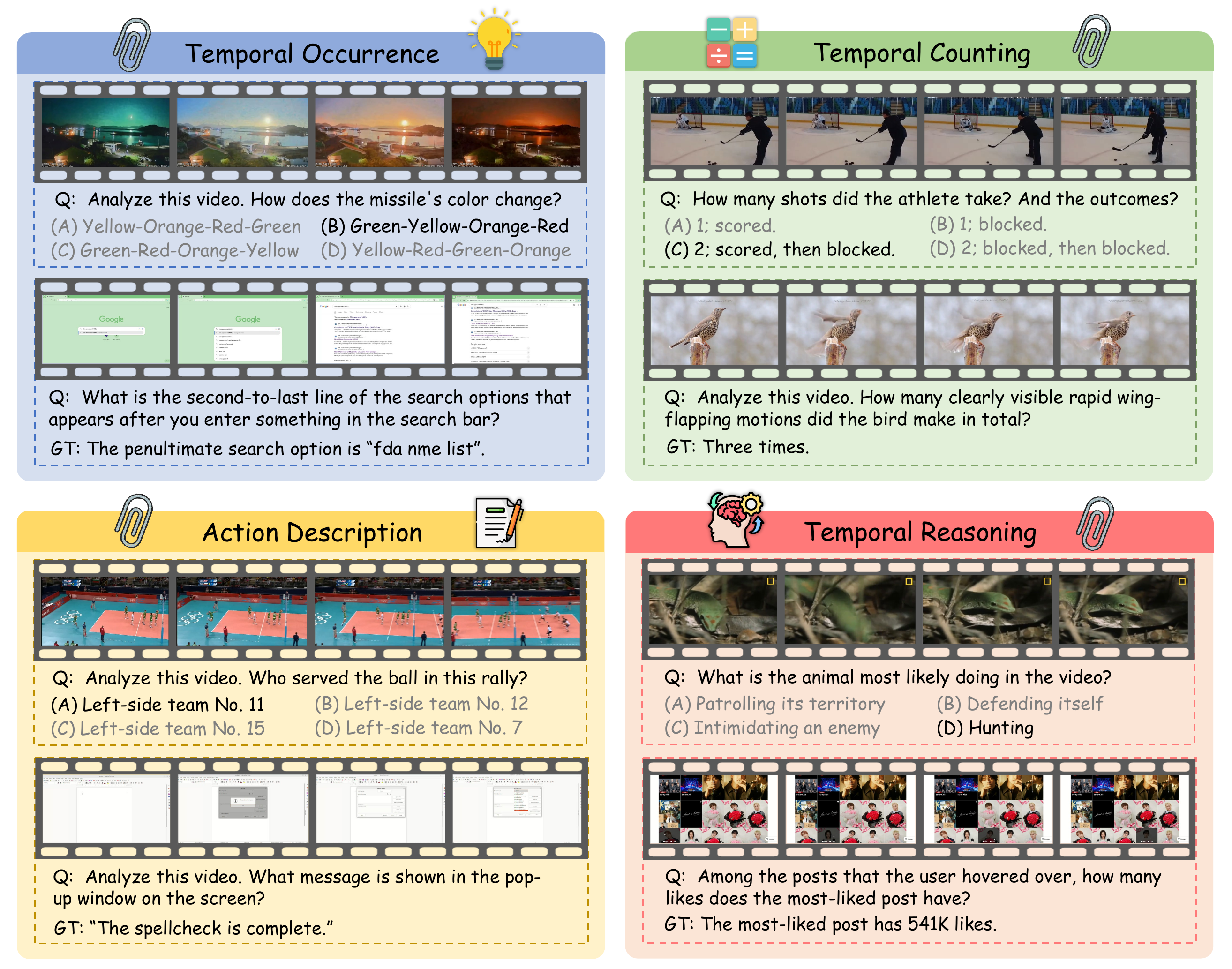}
    \caption{Examples of the four task types in \ourmethod. For each type, we show two questions: one closed-ended and one open-ended.}
    \label{fig:task_examples}
\end{figure}

We organize questions into four task types: Temporal Occurrence, Temporal Counting, Action Description, and Temporal Reasoning, as shown in Fig.~\ref{fig:task_examples}. 
These types decompose temporal fidelity into progressively richer capabilities: noticing brief events, quantifying transient changes, characterizing event dynamics, and reasoning over event-induced state changes. Task formulations and statistics are provided in Appendix~\ref{appendix_dataSta}.

\textbf{Temporal Occurrence (TO).}
Temporal Occurrence assesses a model's ability to identify momentary events or state transitions. Questions in this category probe the presence, absence, or immediate change of a transient visual event. Typical examples include detecting a falling object, a sudden animal movement, a post-operation GUI update, or an unexpected physical anomaly. This task tests the most basic requirement of temporal fidelity: the model must notice the critical moment.

\textbf{Temporal Counting (TC).}
Temporal Counting assesses a model's ability to count transient objects, actions, or event occurrences within a brief visual sequence. Distinct from static object counting, this task necessitates tracking dynamic changes over time while preventing omissions or redundant counts. Typical queries involve quantifying specific actions, state changes, or motions within a critical timeframe. Notably, this task is highly sensitive to sparse sampling, since the omission of even a single brief event can fundamentally alter the final prediction.

\textbf{Action Description (AD).}
Action Description evaluates whether a model can characterize the dynamics of a momentary event. Rather than only determining that an event occurs, the model must describe how it unfolds, including its direction, trajectory, target, interaction, motion pattern, or state transition. For example, a question may ask which direction an object moves after impact, which player performs an action, what interaction happens between two agents, or what interface state changes after a brief operation. This task tests whether the model can accurately report the event-specific dynamic details needed to answer the question.

\textbf{Temporal Reasoning (TR).}
Temporal Reasoning evaluates whether a model can integrate the pre-event state, momentary event, and post-event state to infer a consequence. Questions in this category require more than event recognition. The model must compare states before and after the event, track how the event changes the scene, and infer the resulting condition. For example, after observing several objects initially present and some quickly leaving, falling, or changing state, the model may need to infer how many remain, what state the scene reaches, or why a subsequent change occurs.

\subsection{Evaluation Protocol and Scoring}
\label{sec:answer_format_scoring}

\ourmethod{} contains both multiple-choice and open-ended questions. 
Each question is converted into a binary correctness label according to its answer format. The overall benchmark score is computed as the average accuracy over all 1,000 samples. We additionally report macro accuracy with equal task weights in Appendix~\ref{app:task_macro_average}.

\textbf{Shuffle-robust multiple-choice evaluation.}
All multiple-choice questions have exactly one correct option. Each is evaluated under as many shuffled option orders as it has options.
A prediction is counted as correct only when the model consistently selects the semantically correct option across all shuffled orders. 
Formally, for a multiple-choice sample $i$ with ground-truth answer $a_i^\star$, let $\hat{a}_{i,s}$ denote the model prediction under the $s$-th shuffled option order. 
The sample is scored as correct only if
\begin{equation}
    \mathrm{Acc}_{\mathrm{MC}}(i) =
    \textbf{1}\left[
    \bigwedge_{s=1}^{S_i} \hat{a}_{i,s} = a_i^\star
    \right],
\end{equation}
where $S_i$ equals the number of options in sample $i$ (e.g., four evaluations for four options).
This shuffle-robust criterion reduces option-position bias and discourages models from relying on superficial answer ordering. 
When a model outputs the option content instead of the option letter, we map the response to the corresponding option by semantic matching.

\textbf{Open-ended semantic evaluation.}
For open-ended questions, exact string matching is insufficient because semantically equivalent answers may be expressed in different forms. We therefore use an LLM-as-judge protocol to evaluate semantic consistency between the model prediction and the reference answer. The evaluator receives the reference answer, and the model prediction, and outputs a binary correctness label. It is instructed to judge only answer equivalence and not to introduce external knowledge or reinterpret the video content. Vague, incomplete, contradictory, or overly broad answers are marked as incorrect. The full judge prompt and the human validation of judge reliability are provided in Appendix~\ref{appendix_prompt} and~\ref{appendix_judgeValid}.

%% file: sections/experiments.tex
\section{Experiments}

\subsection{Evaluation Setup}

We summarize the evaluated models and video input settings below; Appendix~\ref{app:eval_setup} details the evaluation protocols.

\textbf{Models:}
We evaluate both proprietary and open-source video MLLMs on \ourmethod{} to examine their ability to capture, preserve, and reason over brief, answer-critical visual evidence. 
Proprietary models: Seed-2.0 variants~\citep{seed}, Gemini series~\citep{gemini3}, GPT-5.4~\citep{GPT5}. 
Open-source models: Qwen series~\citep{qwen3_5,qwen3_vl,qwen3.6-27b,qwen3.6-35b-a3b,qwen35_omni}, DeepSeek-v4.1-Flash~\citep{ds41}, InternVL3.5~\citep{internvl35}, Kimi 2.6~\citep{kimi2.6}, and MIMO-v2.5~\citep{mimov25}, LLaVA-Video~\citep{llava}, VideoLLaMA3~\citep{videollama3}, GLM~\citep{glm45v}, Gemma~\citep{gemma4}, Keye-VL~\citep{keye}, and VITA~\citep{vita1.5} models. 

\textbf{Video input processing:} For video-native models, we directly feed the original video clip into the model whenever its inference interface supports video input. For models that do not support video input, we convert each video into a sequence of sampled frames and provide them as multi-image input. 
Most image-based models are evaluated with 1 FPS sampling. 
If the sampled frames exceed the specified frame budget, we uniformly subsample them across the full video duration.

\input{tables/main_result}

\subsection{Main Experimental Results}

\noindent\textbf{Overall Performance.}

Table~\ref{tab:main_results} summarizes the main results on \ourmethod. Overall, current MLLMs show substantial difficulty in momentary visual event understanding. Among proprietary models, Seed-2.0-Pro achieves the best overall accuracy of 39.6\%, followed by Seed-2.0-Lite with 31.3\% and Seed-2.0-Mini with 27.8\%. Other proprietary models, including Gemini-series models and GPT-5.4, obtain accuracies between 20.4\% and 26.9\%. Seed's ordering changes under a common 1 FPS, 64-frame cap (Appendix~\ref{app:seed_common_budget}), indicating sensitivity to input configuration. Appendix~\ref{8fps_result} details 8 FPS results and model-specific frame budgets.

For open-source models, Kimi-2.6 achieves the strongest overall performance with 24.9\%, followed by Qwen3.5-27B with 24.1\%, Qwen3.5-397B-A17B with 22.8\%, DeepSeek-v4.1-Flash with 22.6\%, and Qwen3.6-27B with 22.5\%. MIMO-v2.5 obtains 22.1\%, while several Qwen variants obtain around 20\%--21\%. Many video-oriented open-source models, including Qwen3-VL, InternVL3.5, LLaVA-Video, VideoLLaMA3, and VITA variants, remain below 18\%. This suggests that model scale or accepting video input alone does not guarantee reliable momentary visual event understanding. Removing video input lowers accuracy by 9.40--20.90 points across six models (Appendix~\ref{app:text_only}), confirming the contribution of visual evidence.

\input{tables/ablation_frameRate}

\noindent\textbf{Task-wise Observations.}
Across task types, Action Description generally achieves the highest average performance, whereas Temporal Counting is consistently the most challenging; results on Temporal Occurrence and Temporal Reasoning vary more across models. For example, Seed-2.0-Pro achieves 50.37\% on TO and 47.08\% on AD, but only 31.14\% on TC. Similar trends appear in open-source models: Kimi-2.6 reaches 36.69\% on AD and 30.59\% on TR, but 15.89\% on TC, while Qwen3.5-27B obtains 32.47\% on AD and 19.07\% on TC. These results suggest that models are relatively effective at detecting and describing brief events, but struggle to count repeated transient actions. Temporal Reasoning also remains difficult; its questions require relating states before, during, and after a brief event.

\noindent\textbf{Category-wise Observations.}
Performance also varies substantially across visual domains. Models generally perform better on Nature and Animal videos: Seed-2.0-Pro reaches 69.44\% on Nature and 55.00\% on Animal, while Gemini-3.1-Flash-Lite achieves the highest proprietary-model Animal accuracy of 57.00\%. Among open-source models, MIMO-v2.5 obtains the best Nature accuracy of 52.78\%, and Qwen3.5-27B achieves the best Animal accuracy of 53.00\%. Visually salient motion may contribute to these results. AIGC, Industry, Games, and Human videos show lower performance; synthetic artifacts, subtle movements, fast camera motion, and complex interactions are possible contributors, but their effects are not isolated here. Additional results are provided in Appendix~\ref{appendix_result1}.

\subsection{Frame-rate Analysis}

We evaluate Gemini-3.1-Pro, Gemini-3-Flash, Qwen3.5-27B, and Qwen3.5-397B-A17B at 1, 5, 8, and 16 FPS to study temporal sampling density. As shown in Table~\ref{tab:frame_rate_ablation}, denser sampling generally improves performance, but gains are model-dependent and not strictly monotonic. Gemini-3.1-Pro improves from 26.9\% at 1 FPS to 38.3\% at 5 FPS, then obtains 36.3\% and 36.6\% at 8 and 16 FPS. Gemini-3-Flash rises from 26.9\% to 37.0\% at 8 FPS, but drops to 34.9\% at 16 FPS. Qwen3.5-27B gains gradually from 24.1\% to 25.7\%, while Qwen3.5-397B-A17B rises from 22.8\% to 26.4\% at 5 FPS and 27.1\% at 16 FPS. For frame-sequence models, the cap also rises from 64 to 128 frames above 1 FPS. Paired short-video results without frame-budget subsampling likewise show model-dependent changes (Appendix~\ref{app:short_video_pairing}).

A timestamp within an annotated event interval does not establish that the sampled frames show its direction, repetitions, or relevant state changes. We therefore supplement the coverage audit with a human review of evidence sufficiency on 300 samples (Appendix~\ref{app:event_coverage}, Table~\ref{tab:event_coverage}). Two annotators assess the sampled frames against the question and reference answer. At 8 FPS, 278--291 samples per open-weight model are judged to contain sufficient evidence, yet accuracy on these samples remains only 21.2\%--31.6\%. Missing visual evidence therefore does not explain all errors.

\newpage
\subsection{Duration Analysis}

\begin{wrapfigure}{r}{0.65\linewidth}
    \vspace{-10pt}
    \centering
    \includegraphics[width=\linewidth]{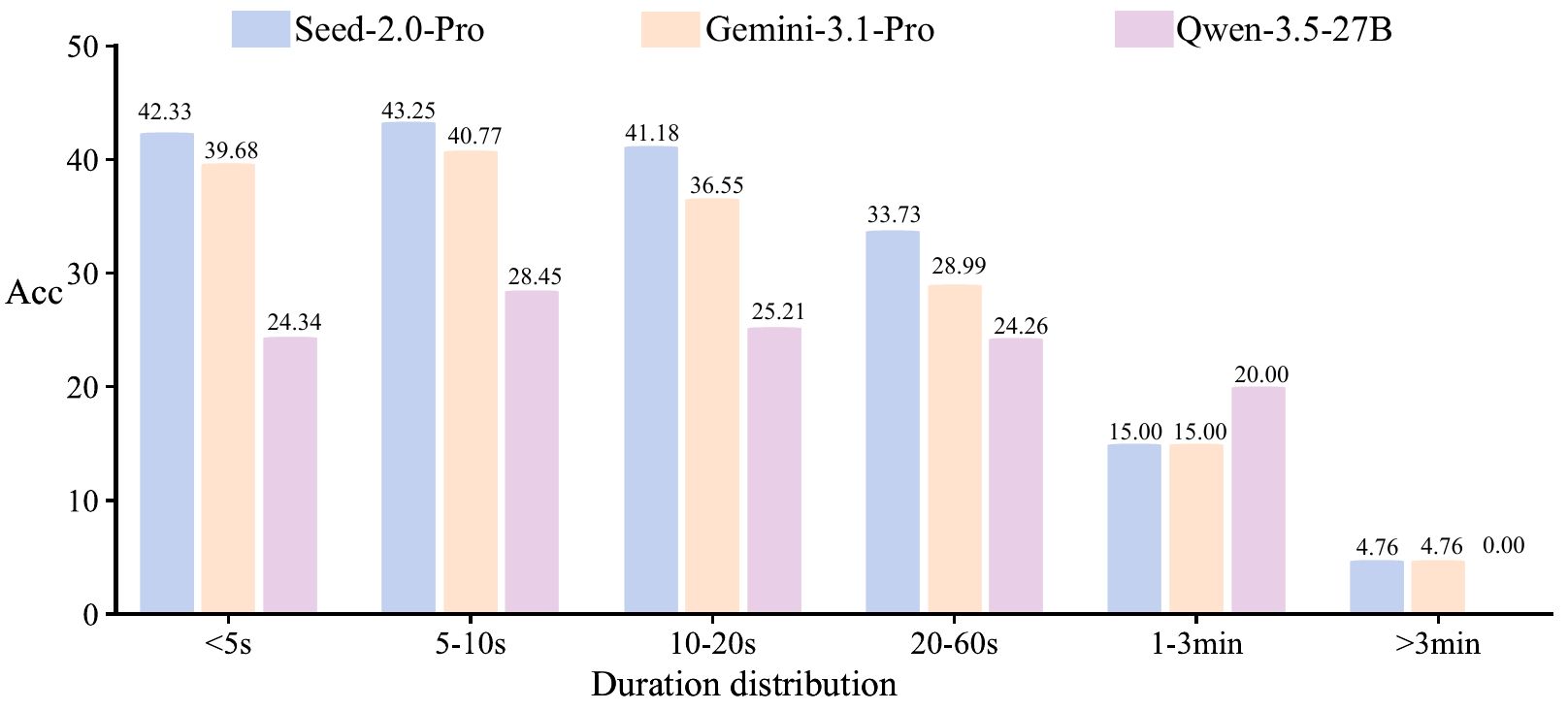}
    \caption{Accuracy of three representative models across video-duration buckets.}
    \label{fig:duration_analysis}
\end{wrapfigure}

We analyze Seed-2.0-Pro, Gemini-3.1-Pro, and Qwen3.5-27B across six duration bins: $\leq$5s, 5--10s, 10--20s, 20--60s, 1--3min, and $\geq$3min. The latter two models use 8 FPS inputs.

As shown in Figure~\ref{fig:duration_analysis}, performance generally decreases as videos become longer. Seed-2.0-Pro maintains over 41\% accuracy below 20 seconds, but drops to 33.73\% at \mbox{20--60s} and 4.76\% beyond 3 minutes. Gemini-3.1-Pro similarly decreases from 40.77\% at 5--10s to 28.99\% at 20--60s and 4.76\% in the longest bin; Qwen3.5-27B reaches 0.00\% in that bin. This pattern is consistent with greater difficulty locating and using brief evidence within longer temporal contexts. The 5--10s bucket performs comparably to or better than the $\leq$5s bucket; surrounding context is one possible explanation, alongside differences in sample composition. Since the two longest buckets contain relatively few samples, these results should be interpreted as indicative trends. Appendix~\ref{app:event_localization} further links successful event localization to higher QA accuracy; Appendix~\ref{appendix_result2} reports duration-wise results for all models.

\subsection{Effect of Sampling Strategies}
\label{sec:sampling_strategies}

We compare Uniform with CLIP-based adaptations of AKS~\citep{aks2025} and A.I.R.~\citep{air2026} on 1,000 questions using 64 ordered frames. Neither consistently improves accuracy (Table~\ref{tab:sampling_strategies}).

\input{tables/sampling_strategies}

AKS densifies some events but lowers overall coverage from 582/609 to 567/609; A.I.R. covers 559--562 events despite iterative revisiting. Thus, selecting relevant frames can still miss necessary evidence. Even with all source frames retained on 43 questions, Qwen, Intern, and Gemma answer only 9, 5, and 13 correctly. Reliable understanding therefore requires both event coverage and interpretation of brief visual details, motivating observation strategies tailored to answer-critical events.

\subsection{Temporal Order and Critical-event Ablations}
\label{sec:temporal_ablations}

We shuffle the same input frames on all 1,000 questions and remove frames belonging to answer-critical events on the 300-question annotated subset. Table~\ref{tab:temporal_ablations} reports each intervention against its ordered baseline. Shuffling reduces overall accuracy by 7.90, 2.40, and 5.20 percentage points for Qwen, Intern, and Gemma, respectively; event removal produces larger drops of 15.00, 7.67, and 10.67 points.

\input{tables/temporal_ablation}

The effects differ by task. TO remains stable or improves under shuffling, consistent with its focus on whether a critical event is visible. Removing that evidence reduces TO accuracy in all three models. TC, AD, and TR decrease under both interventions: counting requires distinguishing repeated occurrences, description uses event dynamics, and reasoning relates events to state changes. These complementary responses support the benchmark's sensitivity to event visibility and temporal organization.

%% file: tables/main_result.tex
\begin{table*}[t]
\centering
\small
\caption{
Main results on \textbf{\ourmethod}.
\textbf{Bold} and \underline{underlined} values mark the best and second-best within each model group. ``Default'' denotes provider-managed video preprocessing. Human reference aggregates 300 responses to 200 sampled questions (Appendix~\ref{appendix_human_valid}).
}
\setlength{\tabcolsep}{3.5pt}
\resizebox{\textwidth}{!}{
\begin{tabular}{lccccccccccccc}
\toprule
\multirow{2}{*}{\textbf{Model}} 
& \textbf{Input}
& \multicolumn{4}{c}{\textbf{Task Type(\%)}}
& \multicolumn{7}{c}{\textbf{Video Category(\%)}}
& \multirow{2}{*}{\textbf{Overall(\%)}} \\
\cmidrule(lr){3-6} \cmidrule(lr){7-13}
& \textbf{Settings}
& \textbf{TO} & \textbf{TC} & \textbf{AD} & \textbf{TR} 
& \textbf{AIGC} & \textbf{GUI} & \textbf{Nature} & \textbf{Industry} & \textbf{Games} & \textbf{Human} & \textbf{Animal} & \\
\midrule
Human reference & video & 81.08 & 82.39 & 88.35 & 83.33 & 66.67 & 82.81 & 86.96 & 92.50 & 66.67 & 92.21 & 93.33 & 84.33 \\
\midrule
\rowcolor{gray!15} \multicolumn{14}{c}{\textbf{Proprietary Models}} \\
\midrule
Seed-2.0-Pro~\cite{seed} 
& default & \textbf{50.37} & \textbf{31.14} & \textbf{47.08} & \textbf{42.35} & \textbf{45.57} & \textbf{33.94} & \textbf{69.44} & \textbf{41.32} & \textbf{39.37} & \textbf{27.20} & \underline{55.00} & \textbf{39.6} \\

Seed-2.0-Lite~\cite{seed} 
& default & 34.81 & \underline{25.64} & \underline{36.04} & \underline{40.00} & 22.78 & \underline{27.98} & 51.39 & \underline{30.58} & \underline{25.62} & \underline{26.80} & 52.00 & \underline{31.3} \\

Seed-2.0-Mini~\cite{seed} 
& default & 32.59 & 22.88 & 33.77 & 25.88 & 24.05 & 22.94 & \underline{58.33} & 29.75 & 22.50 & 20.80 & 43.00 & 27.8 \\

Gemini-3.1-Pro~\cite{gemini3}
& 1fps & 32.59 & 19.70 & 33.44 & 34.12 & \underline{27.85} & 27.52 & 44.44 & 18.18 & 24.37 & 19.20 & 46.00 & 26.9 \\

Gemini-3-Flash~\cite{gemini3} 
& 1fps & \underline{41.48} & 18.43 & 31.82 & 32.94 & \textbf{45.57} & 20.64 & 43.06 & 15.70 & 20.62 & 23.20 & 47.00 & 26.9 \\

Gemini-3.1-Flash-Lite~\cite{gemini3} 
& 1fps & 34.07 & 18.64 & 30.52 & 27.06 & \underline{27.85} & 22.02 & 41.67 & 22.31 & 20.00 & 14.00 & \textbf{57.00} & 25.1 \\

GPT-5.4~\cite{GPT5} 
& 1fps & 21.48 & 12.08 & 30.19 & 29.41 & 6.33 & 23.85 & 40.28 & 4.96 & 18.75 & 18.00 & 37.00 & 20.4 \\

\midrule
\rowcolor{gray!15} \multicolumn{14}{c}{\textbf{Open-source Models}} \\
\midrule
Kimi-2.6~\cite{kimi2.6} 
& 1fps & 25.93 & 15.89 & \textbf{36.69} & \underline{30.59} & 15.19 & \textbf{37.16} & 41.67 & 12.40 & 16.25 & \textbf{20.40} & 34.00 & \textbf{24.9} \\

Qwen3.5-27B~\cite{qwen3_5} 
& 1fps & 20.74 & \textbf{19.07} & 32.47 & 27.06 & 11.39 & 21.56 & 41.67 & 18.18 & \underline{18.75} & \underline{20.00} & \textbf{53.00} & \underline{24.1} \\

Qwen3.5-397B-A17B~\cite{qwen3_5} 
& 1fps & \underline{27.41} & 14.83 & \underline{32.79} & 23.53 & 16.46 & 23.39 & \underline{47.22} & 18.18 & 17.50 & 14.80 & 43.00 & 22.8 \\

DeepSeek-v4.1-Flash~\cite{ds41}
& 1fps & \textbf{30.37} & 14.62 & 28.90 & \textbf{31.76} & \textbf{27.85} & \underline{32.57} & 41.67 & 9.92 & 12.50 & 17.20 & 28.00 & 22.6 \\

Qwen3.6-27B~\cite{qwen3.6-27b} 
& 1fps & 19.26 & \underline{17.16} & 31.82 & 23.53 & 10.13 & 26.61 & 41.67 & 19.01 & 16.25 & 14.40 & \underline{44.00} & 22.5 \\

MIMO-v2.5~\cite{mimov25} 
& default & 18.52 & \underline{17.16} & 30.84 & 23.53 & 2.53 & 24.31 & \textbf{52.78} & 19.01 & \textbf{19.38} & 14.80 & 37.00 & 22.1 \\

Qwen3.5-122B-A10B~\cite{qwen3_5} 
& 1fps & 25.19 & 14.19 & 27.60 & 23.53 & \underline{18.99} & 22.02 & 40.28 & 17.36 & 13.75 & 14.40 & 35.00 & 20.6 \\

Qwen3.6-35B-A3B~\cite{qwen3.6-35b-a3b} 
& 1fps & 17.04 & 14.41 & 29.22 & 24.71 & 3.80 & 18.35 & 41.67 & \textbf{22.31} & 15.62 & 15.20 & 39.00 & 20.2 \\

Qwen3.5-35B-A3B~\cite{qwen3_5} 
& 1fps & 20.00 & 16.10 & 26.30 & 18.82 & 8.86 & 19.27 & 45.83 & \underline{19.83} & 13.75 & 12.80 & 40.00 & 20.0 \\

Gemma-4-31B~\cite{gemma4} 
& 1fps & \textbf{30.37} & 13.56 & 24.35 & 22.35 & \textbf{27.85} & 11.93 & 41.67 & 17.36 & 15.62 & 15.60 & 36.00 & 19.9 \\

Qwen3.5-9B~\cite{qwen3_5} 
& 1fps & 18.52 & 13.35 & 26.62 & 18.82 & 10.13 & 17.43 & 40.28 & 12.40 & 11.87 & 14.80 & 40.00 & 18.6 \\

Qwen3.5-4B~\cite{qwen3_5} 
& 1fps & 19.26 & 13.77 & 21.43 & 17.65 & 8.86 & 16.97 & 38.89 & 13.22 & 15.00 & 11.20 & 32.00 & 17.2 \\

Qwen3-VL-235B-A22B~\cite{qwen3_vl} 
& 1fps & 15.56 & 12.08 & 23.38 & 23.53 & 5.06 & 16.97 & 38.89 & 10.74 & 10.62 & 14.00 & 36.00 & 17.0 \\

InternVL3.5-30B-A3B~\cite{internvl35}
& 1fps & 12.59 & 16.10 & 21.75 & 11.76 & 3.80 & 10.55 & 38.89 & 15.70 & 15.00 & 18.40 & 27.00 & 17.0 \\

Qwen3-VL-30B-A3B~\cite{qwen3_vl} 
& 1fps & 18.52 & 13.14 & 20.45 & 20.00 & 8.86 & 14.22 & 43.06 & 14.05 & 15.62 & 9.60 & 32.00 & 16.7 \\

InternVL3.5-8B~\cite{internvl35} 
& 1fps & 15.56 & 15.04 & 20.13 & 15.29 & 3.80 & 11.93 & 38.89 & 13.22 & 13.75 & 14.00 & 37.00 & 16.7 \\

LLaVA-Video-72B~\cite{llava}  
& 1fps & 13.33 & 15.04 & 18.83 & 17.65 & 0.00 & 8.26 & 34.72 & \underline{19.83} & 16.25 & 15.20 & 31.00 & 16.2 \\

InternVL3.5-241B-A28B~\cite{internvl35} 
& 1fps & 19.26 & 11.02 & 20.45 & 18.82 & 1.27 & 12.39 & 40.28 & 9.09 & 11.87 & 13.60 & 36.00 & 15.7 \\

Keye-VL-1.5-8B~\cite{keye} 
& 1fps & 14.81 & 12.92 & 20.45 & 14.12 & 0.00 & 10.09 & 29.17 & 17.36 & 14.37 & 14.40 & 33.00 & 15.6 \\

InternVL3.5-4B~\cite{internvl35} 
& 1fps & 14.81 & 13.14 & 17.86 & 7.06 & 3.80 & 10.55 & 29.17 & 14.05 & 13.75 & 10.80 & 30.00 & 14.3 \\

GLM-4.6V~\cite{glm45v} 
& 1fps & 15.56 & 8.69 & 19.81 & 21.18 & 5.06 & 10.55 & 31.94 & 7.44 & 11.25 & 12.40 & 33.00 & 14.1 \\

Qwen3-VL-8B~\cite{qwen3_vl} 
& 1fps & 15.56 & 10.59 & 17.86 & 16.47 & 8.86 & 13.30 & 41.67 & 9.92 & 6.88 & 8.80 & 29.00 & 14.0 \\

Qwen3-VL-4B~\cite{qwen3_vl} 
& 1fps & 14.81 & 8.69 & 19.48 & 21.18 & 1.27 & 14.22 & 41.67 & 7.44 & 9.38 & 9.20 & 30.00 & 13.9 \\

LLaVA-Video-7B~\cite{llava} 
& 1fps & 11.11 & 13.14 & 13.96 & 7.06 & 0.00 & 8.72 & 23.61 & 8.26 & 15.62 & 12.80 & 23.00 & 12.6 \\

VideoLLaMA3-7B~\cite{videollama3} 
& 1fps & 7.52 & 13.69 & 11.72 & 8.33 & 0.00 & 10.44 & 25.35 & 4.96 & 10.14 & 15.64 & 15.00 & 11.8 \\

GLM-4.6V-Flash~\cite{glm45v} 
& 1fps & 11.11 & 7.20 & 14.29 & 20.00 & 1.27 & 8.72 & 43.06 & 13.22 & 10.00 & 6.00 & 12.00 & 11.0 \\

VITA-1.5~\cite{vita1.5} 
& 1fps & 19.26 & 8.49 & 13.42 & 10.59 & 0.00 & 5.31 & 25.00 & 8.26 & 11.25 & 8.80 & 26.00 & 10.6 \\
\bottomrule
\end{tabular}
}
\label{tab:main_results}
\end{table*}

%% file: tables/ablation_frameRate.tex
\begin{table}[t]
\centering
\footnotesize %
\caption{
Frame-rate ablation on \textbf{\ourmethod} at 1/5/8/16 FPS. Frame-sequence inputs use at most 64 frames at 1 FPS and 128 at 5/8/16 FPS. Bold marks each model's best result per metric.
}
\label{tab:frame_rate_ablation}

\begin{tabular}{lcccccc}
\toprule
\textbf{Model}
& \textbf{Frame Rate}
& \textbf{TO(\%)}
& \textbf{TC(\%)}
& \textbf{AD(\%)}
& \textbf{TR(\%)}
& \textbf{Overall(\%)} \\
\midrule

\multirow{4}{*}{
  Gemini-3.1-Pro~\cite{gemini3}
}
& 1 FPS
& 32.59
& 19.70
& 33.44
& 34.12
& 26.9 \\

& 5 FPS
& \textbf{42.22}
& \textbf{30.08}
& \textbf{48.05}
& 42.35
& \textbf{38.3} \\

& 8 FPS
& 40.00
& 27.75
& 45.45
& 44.71
& 36.3 \\

& 16 FPS
& 40.00
& 27.97
& 45.78
& \textbf{45.88}
& 36.6 \\

\midrule

\multirow{4}{*}{
  Gemini-3-Flash~\cite{gemini3}
}
& 1 FPS
& \textbf{41.48}
& 18.43
& 31.82
& 32.94
& 26.9 \\

& 5 FPS
& 34.07
& 27.75
& \textbf{49.35}
& \textbf{38.82}
& 36.2 \\

& 8 FPS
& 40.00
& \textbf{27.97}
& 49.03
& \textbf{38.82}
& \textbf{37.0} \\

& 16 FPS
& 39.26
& 25.42
& 46.43
& \textbf{38.82}
& 34.9 \\

\midrule

\multirow{4}{*}{
  Qwen3.5-27B~\cite{qwen3_5}
}
& 1 FPS
& 20.74
& 19.07
& 32.47
& \textbf{27.06}
& 24.1 \\

& 5 FPS
& 17.78
& 18.64
& 36.69
& \textbf{27.06}
& 24.8 \\

& 8 FPS
& 22.22
& \textbf{19.92}
& 36.36
& 20.00
& 25.3 \\

& 16 FPS
& \textbf{24.44}
& 18.01
& \textbf{38.64}
& 23.53
& \textbf{25.7} \\

\midrule

\multirow{4}{*}{
  Qwen3.5-397B-A17B~\cite{qwen3_5}
}
& 1 FPS
& 27.41
& 14.83
& 32.79
& 23.53
& 22.8 \\

& 5 FPS
& \textbf{29.63}
& 18.43
& \textbf{37.66}
& 24.71
& 26.4 \\

& 8 FPS
& \textbf{29.63}
& 17.16
& 36.69
& 25.88
& 25.6 \\

& 16 FPS
& 27.41
& \textbf{19.70}
& \textbf{37.66}
& \textbf{29.41}
& \textbf{27.1} \\

\bottomrule
\end{tabular}
\end{table}

%% file: tables/sampling_strategies.tex
\begin{table}[htbp]
\centering
\small
\caption{Accuracy (\%) under three sampling strategies on 1,000 questions. Each strategy supplies 64 ordered frames for the final answer.}
\label{tab:sampling_strategies}
\begin{tabular}{lrrr}
\toprule
Backbone & Uniform & AKS-CLIP & A.I.R.-CLIP \\
\midrule
Qwen3.6-27B & 21.4 & 21.2 & 20.5 \\
InternVL3.5-30B-A3B & 10.4 & 10.3 & 9.7 \\
Gemma-4-31B & 24.0 & 24.9 & 25.0 \\
\bottomrule
\end{tabular}
\end{table}

%% file: tables/temporal_ablation.tex
\begin{table}[htbp]
\centering
\small
\setlength{\tabcolsep}{3pt}
\caption{Temporal ablations: accuracy (\%) and overall change from the paired ordered baseline (percentage points). Shuffling uses 1,000 questions (TO/TC/AD/TR: 135/472/308/85); event removal uses 300 (40/142/92/26).}
\label{tab:temporal_ablations}
\begin{tabular}{llrrrrrr}
\toprule
Model & Input ($n$) & TO & TC & AD & TR & Overall & $\Delta$ \\
\midrule
\multirow{4}{*}{\shortstack[l]{Qwen3.6-\\27B}}
& Ordered (1,000) & 28.89 & 21.40 & 34.74 & 30.59 & 27.30 & -- \\
& Shuffled (1,000) & 29.63 & 11.02 & 26.30 & 24.71 & 19.40 & $-7.90$ \\
& Ordered (300) & 17.50 & 28.87 & 38.04 & 34.62 & 30.67 & -- \\
& Event removed (300) & 5.00 & 13.38 & 23.91 & 15.39 & 15.67 & $-15.00$ \\
\midrule
\multirow{4}{*}{\shortstack[l]{InternVL3.5-\\30B-A3B}}
& Ordered (1,000) & 17.78 & 14.19 & 23.70 & 16.47 & 17.80 & -- \\
& Shuffled (1,000) & 18.52 & 10.81 & 22.08 & 11.76 & 15.40 & $-2.40$ \\
& Ordered (300) & 17.50 & 17.61 & 25.00 & 26.92 & 20.67 & -- \\
& Event removed (300) & 7.50 & 11.27 & 17.39 & 15.38 & 13.00 & $-7.67$ \\
\midrule
\multirow{4}{*}{\shortstack[l]{Gemma-4-\\31B}}
& Ordered (1,000) & 32.59 & 17.16 & 31.17 & 27.06 & 24.40 & -- \\
& Shuffled (1,000) & 34.07 & 8.05 & 28.57 & 23.53 & 19.20 & $-5.20$ \\
& Ordered (300) & 25.00 & 19.01 & 34.78 & 26.92 & 25.33 & -- \\
& Event removed (300) & 10.00 & 11.97 & 20.65 & 15.38 & 14.67 & $-10.67$ \\
\bottomrule
\end{tabular}
\end{table}

%% file: sections/conclusion.tex
\section{Conclusion}

We introduce \ourmethod{}, a benchmark of 1,000 human-verified video-QA pairs spanning seven domains and four tasks. Tests on 34 MLLMs reveal difficulty using brief, answer-critical evidence, especially for temporal counting. Denser sampling helps some models, yet errors persist in the audited subset even when human review judges the sampled evidence sufficient to answer. Missing visual evidence therefore does not explain all failures. Lower accuracy on longer videos also suggests localization challenges. These findings motivate video models that better capture, preserve, and use brief visual events.

%% file: sections/appendix.tex
\setcounter{section}{0}
\section{Benchmark Data Sources}
\label{appendix_dsource}

\ourmethod{} is constructed from diverse video sources to cover momentary visual event across both real-world and virtual scenarios. The data sources include existing open-source video benchmarks, self-collected screen recordings, self-recorded real-world videos, AI-generated videos, and publicly available online videos such as YouTube clips. During data collection, we prioritize videos that contain visually clear momentary actions or short-duration state transitions. We exclude videos whose key events are severely blurred, heavily occluded, or mainly dependent on audio, subtitles, or external context. For videos collected from public online sources, we record the source metadata and access information. Raw videos are released only when redistribution is permitted by the corresponding license or when the videos are self-collected; otherwise, we release annotations, metadata, and source links. Table~\ref{tab:data_sources} summarizes the data sources for each fine-grained subcategory in \ourmethod. The table is intended to document the provenance of different video types and improve the transparency of benchmark construction.

\section{Experimental Setup}
\label{app:eval_setup}

We use each model's supported input interface and preprocessing pipeline. Models with native video support are evaluated through their video interfaces; the explicit frame budgets below apply to frame-sequence inputs, not provider-managed video preprocessing. For explicitly sampled inputs, we first extract candidate frames at the specified rate. If their number exceeds the frame budget, we uniformly subsample them across the full video duration rather than truncating the video.

In the main evaluation, frame-sequence models use 1 FPS sampling with at most 64 frames per video, except GPT-5.4, which uses at most 50 frames through the Amazon API. Seed-2.0-Pro, Seed-2.0-Lite, and Seed-2.0-Mini use their native video APIs with default processing. Gemini-3.1-Pro, Gemini-3-Flash, and Gemini-3.1-Flash-Lite use their supported video interfaces with the sampling rate set to 1 FPS. MIMO-v2.5 uses its native video interface with default processing.

Kimi-2.6 and DeepSeek-v4.1-Flash are evaluated through the OpenRouter API. LLaVA-Video-72B, LLaVA-Video-7B, VideoLLaMA3-7B, and VITA-1.5 receive offline-extracted frame sequences. The remaining locally deployed open-source models are served through vLLM, which decodes videos and constructs the sampled frame sequences before forwarding them to the multimodal processor. These explicitly sampled configurations follow the frame budgets specified here.

For the 5, 8, and 16 FPS experiments, frame-sequence models receive at most 128 frames per video; GPT-5.4 retains its 50-frame limit in every evaluated sampling configuration. Native video models continue to use their supported interfaces and provider-specific preprocessing. For short videos, the actual frame count may be below the cap. For longer videos, uniform subsampling can reduce the effective input rate below the nominal candidate sampling rate.

\section{Dataset Statistics}
\label{appendix_dataSta}
This section provides detailed statistics of \textbf{\ourmethod}. The benchmark contains 1,000 videos and 1,000 question-answer pairs, with each video corresponding to one question. We report domain and subcategory coverage, task-type composition, answer-format distribution, video duration, source composition, and an event-level temporal audit. These statistics show that \ourmethod{} covers diverse visual scenarios and evaluates multiple levels of momentary visual event understanding.

\paragraph{Domain and Subcategory Distribution.}
Table~\ref{tab:subcategory_statistics} reports the number of samples in each fine-grained subcategory. \ourmethod{} covers 7 high-level domains and 25 subcategories, including both real-world scenarios, such as nature, industry, human activities, and animal behaviors, and virtual or synthetic scenarios, such as AIGC videos, GUI recordings, and games.

\paragraph{Task-type Distribution.}
Table~\ref{tab:task_statistics} shows the distribution of the four task types in \ourmethod. Temporal Occurrence evaluates whether a momentary event occurs, Temporal Counting focuses on counting transient actions or state changes, Action Description requires describing event dynamics, and Temporal Reasoning requires inferring consequences from pre-event, event, and post-event information.

\paragraph{Answer-format Distribution.}
{\ourmethod} contains both open-ended and multiple-choice questions. We use open-ended questions when the answer is convergent and can be semantically matched with a canonical reference answer. When the answer may be difficult to standardize or could introduce ambiguity in automatic evaluation, we convert the question into a multiple-choice format with manually written distractors. Table~\ref{tab:answer_format_statistics} reports the distribution of answer formats.

\paragraph{Video Duration and Source Composition.}
The video durations range from 1.00 to 560.00 seconds, with a mean of 19.25 seconds, a median of 8.02 seconds, and an interquartile range of 5.05--16.34 seconds. In total, 79.0\% of videos are shorter than 20 seconds, 41 videos are longer than one minute, and 21 videos are longer than three minutes. The duration buckets contain 190 videos shorter than 5 seconds, 363 videos of 5--10 seconds, 237 videos of 10--20 seconds, 169 videos of 20--60 seconds, 20 videos of 1--3 minutes, and 21 videos longer than 3 minutes. Among the 1,000 videos, 158 (15.8\%) are self-recorded or self-collected, including GUI recordings, and 842 (84.2\%) come from existing public video sources.

\paragraph{Event Duration and Temporal Position Audit.}
To quantify the temporal locality of the benchmark events, we conduct a stratified audit of 300 samples covering all domains, subcategories, task types, answer formats, and duration ranges. The audit identifies 609 answer-critical event intervals, including all necessary events for counting and temporal-reasoning questions. The mean and median event durations are 0.703 and 0.540 seconds; 47.9\%, 87.7\%, and 97.9\% of the intervals last no longer than 0.5, 1, and 2 seconds, respectively. A total of 151/300 samples (50.3\%) require two or more necessary events. Event centers are distributed across the video timeline rather than concentrated at a fixed location: 14.1\%, 20.2\%, 22.8\%, 23.6\%, and 19.2\% fall in the 0--20\%, 20--40\%, 40--60\%, 60--80\%, and 80--100\% relative-position bins, respectively. For an annotated interval $\tau_j=[t_{s,j},t_{e,j}]$ in a video of duration $T$, we measure its duration $\Delta_j=t_{e,j}-t_{s,j}$ and relative duration $r_j=\Delta_j/T$. In this audit, 71.4\% of necessary event intervals satisfy $r_j\leq0.10$. These are per-interval proportions, not the union of all necessary intervals in a video. The intervals describe answer-critical visual evidence, which can be shorter than the surrounding action. The duration statistics characterize the audited subset; neither 1 second, 2 seconds, nor 10\% is a hard inclusion threshold. Together with the multi-event counts, they show that models must locate and integrate brief evidence at unknown positions.

\paragraph{Task Formulations and Evaluated Capabilities.}
Given a video $V$ and question $q$, a model predicts an answer $\hat y=f(V,q)$. Let $\mathcal{E}_q=\{\tau_j\}_{j=1}^{m}$ denote the necessary evidence intervals identified for an audited question. These intervals serve the coverage analysis and are not supplied to the model; the full benchmark uses weak temporal grounding during construction rather than dense interval annotations for every sample. The question determines which event, entities, and temporal scope are relevant.

For \textbf{TO}, the answer identifies the presence, absence, or immediate change of a queried event or state. For a yes/no occurrence question, this reduces to $y=\mathbf{1}[e\text{ occurs in }V]$; other TO questions name the observed change. The capability is noticing and identifying brief evidence, rather than counting repetitions. For \textbf{TC}, $y=|\mathcal{O}_q|$, where $\mathcal{O}_q$ is the set of distinct occurrences, entities performing the specified action, or state changes requested by $q$. The capability is temporal tracking and quantification without omissions or double counting. For \textbf{AD}, $y=d_q(\mathcal{E}_q)$ describes the requested visible event attributes, such as actor, direction, trajectory, target, or interaction. The capability is retaining event-specific dynamic details. For \textbf{TR}, $y=g_q(s_{\mathrm{pre}},\mathcal{E}_q,s_{\mathrm{post}})$ is a consequence inferred by relating the initial state, necessary events, and resulting state. The capability is integrating evidence across time: counting observed departures is TC, whereas inferring how many objects remain from the initial count and departures is TR. These are semantic task targets; each can be expressed as either a multiple-choice or an open-ended question under the scoring protocol in Section~\ref{sec:answer_format_scoring}.

\input{tables/data_source}
\input{tables/appendix_dataStatic}

\section{Data Quality Control}
\label{appd:data_quality_control}

\begin{figure*}[t]
    \centering
    \includegraphics[width=\textwidth]{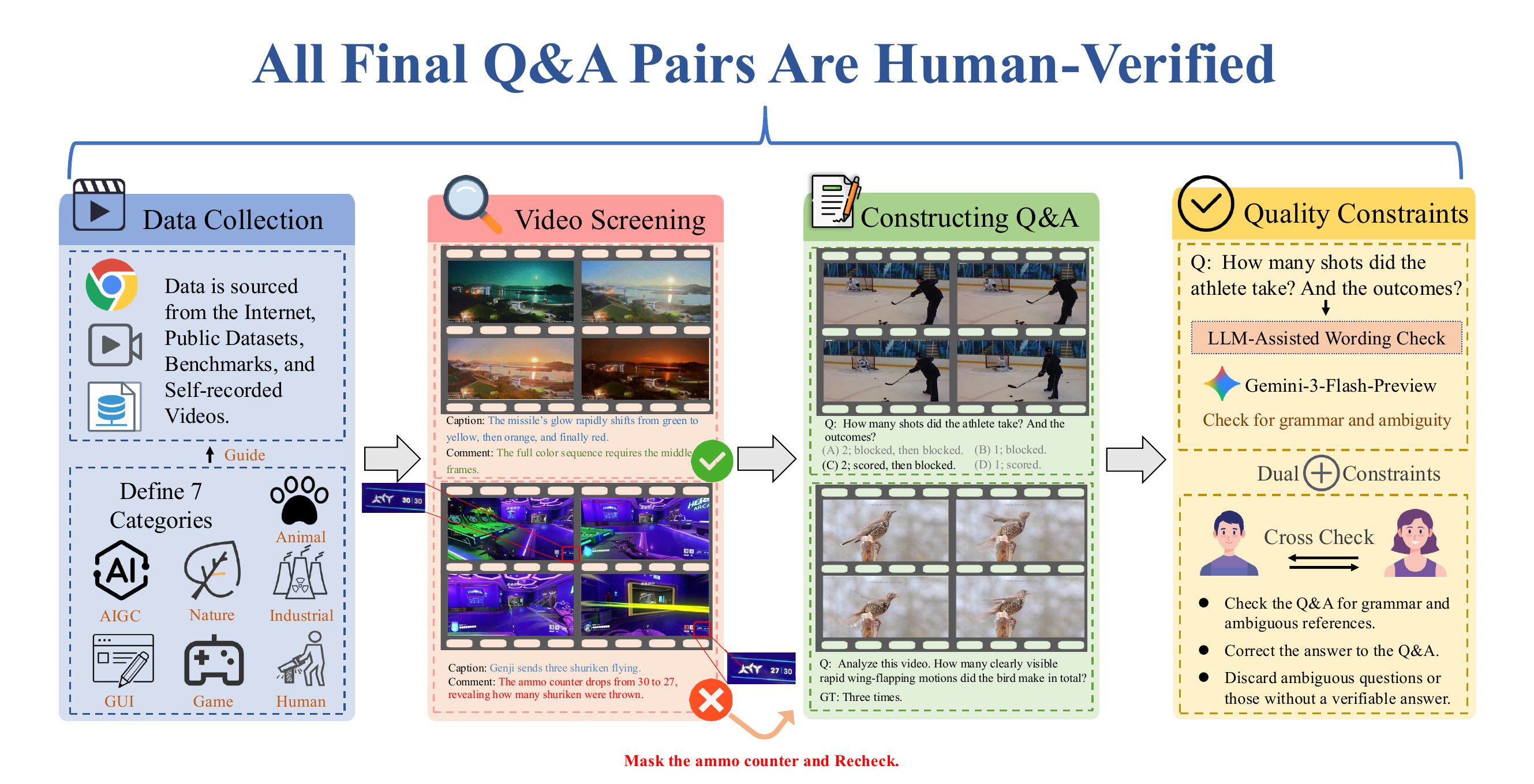}
    \caption{Dataset construction and verification pipeline. All videos follow the same screening criteria: visual evidence must determine a unique answer, and surrounding context must not substitute for the target event. The missile example requires intermediate color stages; the gameplay example illustrates masking an ammunition-counter shortcut and rechecking the sample. All final Q\&A pairs are human-verified, with Gemini-3-Flash-Preview assisting the wording check.}
    \label{fig:dataset_pipeline}
\end{figure*}

We apply the same screening standards to all 1,000 videos, including the 842 videos drawn from existing datasets and other public sources. As illustrated in Fig.~\ref{fig:dataset_pipeline}, annotators first identify a brief, clearly visible action or state change and record its approximate time, involved objects, and relevant before-and-after states. They then construct the question and reference answer around this visual evidence.

Two annotators cross-check every sample against two criteria: the visual evidence determines a unique answer, and the surrounding visual context alone cannot reveal that answer. Specifically, they check whether a viewer who misses the target moment could still infer the answer from the before-and-after states or persistently displayed clues. Samples with such shortcuts are revised or removed. When persistent displayed information can be masked, the masked sample must pass the same checks before retention. The annotators also verify the task and domain labels, answer format, and visual grounding; for multiple-choice questions, they check that exactly one option is correct and the distractors are visually plausible. Gemini-3-Flash-Preview flags potential issues such as unclear wording, missing referents, or ambiguous answers, while humans make the final accept, revise, or remove decision.

Figure~\ref{fig:dataset_pipeline} illustrates both screening outcomes. In the retained missile video, recovering the full color sequence requires observing the intermediate yellow and orange stages between green and red. In the gameplay example, the ammunition counter decreases from 30 to 27, revealing that Genji throws three shuriken even if the throwing moment is missed. The unmasked version therefore fails screening; the counter is masked and the sample is rechecked under the same criteria.

\section{Human Solvability Study}
\label{appendix_human_valid}

To further validate the quality of \ourmethod, we conduct a human solvability study on examples sampled from the benchmark. We initially recruit 20 participants, and each participant answers 20 questions after watching the corresponding videos. Participants do not have access to the reference answers or model predictions. After quality filtering, we retain 15 valid questionnaires, resulting in 300 valid human responses covering 200 distinct questions. We compare each human response with the reference answer to evaluate whether the questions are answerable by human viewers.

\begin{table}[t]
\centering
\small
\caption{
Human solvability study on sampled examples from \ourmethod.
}
\label{tab:human_solvability}
\begin{tabular}{lc}
\toprule
\textbf{Metric} & \textbf{Value} \\
\midrule
Initial participants & 20 \\
Valid participants & 15 \\
Questions per valid participant & 20 \\
Valid human responses & 300 \\
Distinct sampled questions & 200 \\
Overall human accuracy & 84.33\% \\
Participant accuracy range & 70.00\%--95.00\% \\
\bottomrule
\end{tabular}
\end{table}

As shown in Table~\ref{tab:human_solvability}, human participants achieve an overall accuracy of 84.33\% on the valid responses. This result indicates that most sampled questions in \ourmethod{} are answerable by human viewers after watching the corresponding videos. Meanwhile, the accuracy is still far from saturated, suggesting that momentary visual event understanding remains non-trivial even for humans and requires careful observation of brief visual evidence.

Table~\ref{tab:human_breakdown} further aggregates the original responses by task and domain. Human accuracy exceeds 80\% in all four tasks and ranges from 66.67\% to 93.33\% across domains, supporting the solvability of the sampled questions. Seed-2.0-Pro's full-benchmark results are included as a reference; the human results summarize 300 responses to 200 sampled questions.

\begin{table}[htbp]
\centering
\small
\caption{Human accuracy by task and domain, aggregated over 300 valid responses to 200 sampled questions. Seed-2.0-Pro results use all 1,000 benchmark questions under the main evaluation settings. All values are accuracy (\%).}
\label{tab:human_breakdown}
\begin{tabular}{lrrrrr}
\toprule
Evaluation & TO & TC & AD & TR & Overall \\
\midrule
Human reference & 81.08 & 82.39 & 88.35 & 83.33 & 84.33 \\
Seed-2.0-Pro & 50.37 & 31.14 & 47.08 & 42.35 & 39.60 \\
\bottomrule
\end{tabular}

\medskip
\begin{tabular}{lrrrrrrr}
\toprule
Evaluation & AIGC & GUI & Nature & Industry & Games & \shortstack{Human\\activities} & Animal \\
\midrule
Human reference & 66.67 & 82.81 & 86.96 & 92.50 & 66.67 & 92.21 & 93.33 \\
Seed-2.0-Pro & 45.57 & 33.94 & 69.44 & 41.32 & 39.37 & 27.20 & 55.00 \\
\bottomrule
\end{tabular}
\end{table}

\section{LLM-as-Judge Validation}
\label{appendix_judgeValid}

To validate the reliability of our LLM-as-Judge protocol for open-ended questions, we conduct a human validation study on the same set of 764 open-ended examples in \ourmethod. Specifically, we use the predictions from Gemini-3-Flash as the validation set and ask three human annotators to independently judge whether each model prediction is semantically consistent with the reference answer. During this process, annotators only see the model answer and the reference answer; they do not have access to the video, model identity, or the LLM-as-Judge decision. We treat each human annotator's judgment as the reference label and evaluate the LLM-as-Judge decision as a binary classification result, where positive indicates that the model answer is judged as correct or semantically consistent.

Since the validation labels are imbalanced, with incorrect answers forming the majority of samples, raw accuracy alone may not fully reflect judge reliability. We therefore report multiple metrics, including accuracy, F1 score, false positive rate (FPR), and Cohen's kappa. Accuracy measures overall agreement between each human annotator and the LLM-as-Judge. F1 score evaluates consistency on the positive class, i.e., answers judged as semantically correct. FPR is particularly important in our setting because false positives correspond to incorrect model answers being accepted as correct, which may overestimate model performance. Cohen's kappa measures agreement beyond chance and provides a more conservative estimate under imbalanced label distributions.

\begin{table}[t]
\centering
\small
\caption{
Validation of LLM-as-Judge against human annotations on the same 764 open-ended Gemini-3-Flash predictions. Cohen's kappa measures agreement beyond chance.
}
\label{tab:judge_validation}
\begin{tabular}{lccccc}
\toprule
\textbf{Annotator} & \textbf{\#Samples} & \textbf{Acc.(\%)} & \textbf{F1(\%)} & \textbf{FPR(\%)} & \textbf{Kappa} \\
\midrule
Annotator 1 & 764 & 96.73 & 88.89 & 2.14 & 0.870 \\
Annotator 2 & 764 & 97.12 & 90.00 & 1.53 & 0.883 \\
Annotator 3 & 764 & 97.51 & 91.24 & 1.07 & 0.898 \\
\midrule
Average & 764 & 97.12 & 90.04 & 1.58 & 0.884 \\
\bottomrule
\end{tabular}
\end{table}

As shown in Table~\ref{tab:judge_validation}, the LLM-as-Judge shows strong agreement with human annotations on the same 764 open-ended examples, achieving 96.73\%--97.51\% accuracy across the three annotators. The F1 score ranges from 88.89\% to 91.24\%, indicating strong consistency on answers judged as semantically correct. Cohen's kappa ranges from 0.870 to 0.898, showing strong agreement beyond chance even under the imbalanced label distribution.

Importantly, the false positive rate remains very low, ranging from 1.07\% to 2.14\%. This suggests that the judge rarely accepts incorrect model responses as correct, reducing the risk of overestimating model performance. Overall, these results indicate that our LLM-as-Judge protocol is well aligned with human judgments and provides a reliable automatic evaluation procedure for open-ended answers in \ourmethod.

\section{Additional Experimental Results}
\label{appendix_additional}
\subsection{Results of \ourmethod{} under an 8 FPS Input Rate}
\label{8fps_result}

Table~\ref{tab:8fps_result} reports additional evaluations at 8 FPS. Frame-sequence inputs are capped at 128 frames per video, except GPT-5.4, which remains capped at 50; native video models use their supported interfaces. Candidate frames exceeding the applicable cap are uniformly subsampled across the full video duration. Thus, 8 FPS is the candidate sampling rate, and the effective input rate can be lower for longer videos. Gray rows show these additional results; non-gray rows retain the main evaluation settings. Bold and underlined values denote the best and second-best results within each model group, respectively.

\input{tables/result_8fps}

The updated 8 FPS results show model-dependent gains. Gemini-3-Flash and Gemini-3.1-Pro reach 37.0\% and 36.3\% overall accuracy, respectively. Among open-source models, DeepSeek-v4.1-Flash achieves 29.0\%, followed by Qwen3.5-122B-A10B (27.4\%), Qwen3.6-27B (27.3\%), Qwen3.5-397B-A17B (25.6\%), and Qwen3-VL-235B-A22B (25.5\%). Improvements are not universal: Keye-VL-1.5-8B decreases from 15.6\% to 13.8\%, and VideoLLaMA3-7B from 11.8\% to 9.6\%. For frame-sequence models other than GPT-5.4, the frame budget also increases from 64 to 128, so their comparisons with 1 FPS do not isolate sampling rate alone.

However, 8 FPS sampling does not fully solve the challenge of momentary visual event understanding. Even with denser temporal inputs, accuracy remains low on tasks involving event detection, counting, description, and reasoning. These results show that increasing the sampling rate alone is insufficient for reliable performance; they do not identify a specific internal cause of the remaining errors.

\subsection{Event Coverage and Evidence Sufficiency Audit}
\label{app:event_coverage}

We audit 300 stratified samples containing 609 necessary event intervals, including every necessary event for counting and temporal-reasoning questions. The subset is selected independently of model correctness. We reconstruct the final input-frame timestamps under each model's evaluation sampling rules. A timestamp inside an annotated interval indicates an interval hit, but does not establish that the frames provide enough visual evidence to answer the question.

We therefore conduct an additional human review of evidence sufficiency. Two annotators inspect the sampled frames for each evaluation setting together with the question and reference answer. The reference specifies the evidence to check; sufficiency is judged from what is actually visible in the sampled frames. The review checks whether the target event is clearly depicted, all required occurrences are observable, and the action dynamics or before/after states needed for the answer are available. A sample is \emph{Full} when its visible evidence is sufficient to answer, \emph{Partial} when some necessary evidence is visible but incomplete, and \emph{Miss} when the necessary evidence is absent. The revised event evidence coverage is measured over the 609 necessary events; Full, Partial, and Miss partition the 300 samples. These groups reflect reviewed visual evidence, rather than timestamp hits alone.

\begin{table}[htbp]
\centering
\small
\setlength{\tabcolsep}{3pt}
\caption{Human-reviewed event coverage and evidence sufficiency on 300 samples. Event evidence coverage is measured over 609 necessary events. Full, Partial, and Miss reflect sufficient, incomplete, and absent evidence, respectively; cells report sample count / QA accuracy (\%). Overall accuracy uses all 300 samples. FPS denotes the candidate sampling rate, subject to the model-specific frame caps.}
\label{tab:event_coverage}
\begin{tabular}{lrrrrrr}
\toprule
Model & FPS & \shortstack{Event evidence\\coverage (\%)} & \shortstack{Full\\n / Acc.} & \shortstack{Partial\\n / Acc.} & \shortstack{Miss\\n / Acc.} & \shortstack{Overall\\(\%)} \\
\midrule
GPT-5.4 & 1 & 41.7 & 65 / 38.5 & 110 / 11.8 & 125 / 23.2 & 22.3 \\
GPT-5.4 & 8 & 90.0 & 253 / 35.6 & 31 / 12.9 & 16 / 25.0 & 32.7 \\
Qwen3.5-27B & 1 & 44.3 & 69 / 34.8 & 114 / 22.8 & 117 / 23.9 & 26.0 \\
Qwen3.5-27B & 8 & 97.4 & 291 / 27.8 & 7 / 14.3 & 2 / 0.0 & 27.3 \\
Qwen3.6-27B & 1 & 44.3 & 69 / 31.9 & 114 / 23.7 & 117 / 21.4 & 24.7 \\
Qwen3.6-27B & 8 & 97.4 & 291 / 31.6 & 7 / 0.0 & 2 / 0.0 & 30.7 \\
Gemma-4-31B & 1 & 44.3 & 69 / 29.0 & 114 / 14.0 & 117 / 21.4 & 20.3 \\
Gemma-4-31B & 8 & 94.3 & 278 / 25.5 & 17 / 17.6 & 5 / 40.0 & 25.3 \\
InternVL3.5-30B-A3B & 1 & 44.3 & 69 / 30.4 & 114 / 18.4 & 117 / 20.5 & 22.0 \\
InternVL3.5-30B-A3B & 8 & 94.3 & 278 / 21.2 & 17 / 11.8 & 5 / 20.0 & 20.7 \\
\bottomrule
\end{tabular}
\end{table}

As shown in Table~\ref{tab:event_coverage}, reviewed event evidence coverage increases from 41.7\%--44.3\% at 1 FPS to 90.0\%--97.4\% at 8 FPS. The review identifies cases where a timestamp falls within an event interval but the frame does not clearly show the event, omits required repetitions, or fails to preserve the relevant action dynamics or state changes. At 1 FPS, Full-group accuracy exceeds the combined Partial+Miss accuracy for all five models. At 8 FPS, 278--291 of 300 samples are judged Full for the four open-weight models, yet their Full-group accuracy remains only 21.2\%--31.6\%. Thus, substantial errors persist even with human-judged sufficient evidence, and missing visual evidence cannot explain all failures. Partial and Miss accuracies need not be ordered because the groups differ in composition and some are small. This audit characterizes visible input evidence; it does not identify which internal processing stage causes the remaining errors.

\subsection{Paired Frame-rate Results on Short Videos}
\label{app:short_video_pairing}

Table~\ref{tab:short_video_pairing} reports paired 1 FPS and 8 FPS accuracies on short-video subsets where neither setting requires candidate-frame subsampling to meet the frame budget. The comparison reuses existing model predictions. GPT-5.4, Qwen3.6-27B, and Gemma-4-31B improve by 14.72, 7.50, and 6.00 percentage points, respectively; Qwen3.5-27B improves by 0.50 points, while InternVL3.5-30B-A3B decreases by 2.50 points.

\begin{table}[htbp]
\centering
\small
\setlength{\tabcolsep}{6pt}
\caption{Paired frame-rate results on short videos without frame-budget subsampling. Accuracy is reported in \%; changes are in percentage points and computed before rounding. Sample counts refer to the paired subset for each model.}
\label{tab:short_video_pairing}
\begin{tabular}{lrrrr}
\toprule
Model & \shortstack{Paired\\samples} & \shortstack{1 FPS\\accuracy (\%)} & \shortstack{8 FPS\\accuracy (\%)} & \shortstack{Change} \\
\midrule
GPT-5.4 & 299 & 17.06 & 31.77 & $+14.72$ \\
Qwen3.5-27B & 200 & 24.50 & 25.00 & $+0.50$ \\
Qwen3.6-27B & 200 & 24.50 & 32.00 & $+7.50$ \\
Gemma-4-31B & 200 & 19.00 & 25.00 & $+6.00$ \\
InternVL3.5-30B-A3B & 200 & 21.50 & 19.00 & $-2.50$ \\
\bottomrule
\end{tabular}
\end{table}

\subsection{Seed Models under a Common Frame Budget}
\label{app:seed_common_budget}

We evaluate Seed-2.0-Pro, Lite, and Mini on all 1,000 questions using an explicit frame-sequence input. We extract candidate frames at 1 FPS and uniformly subsample them across the full video duration when their number exceeds 64. The selected images and their timestamps are supplied in chronological order. This matches the target sampling rate and frame cap used by most frame-sequence models in the main evaluation. The default native-video results use provider-managed preprocessing; the available logs do not expose the actual processed frame counts or internal selection rules.

\begin{table}[htbp]
\centering
\small
\caption{Seed models with default native-video input and a common 1 FPS, at-most-64-frame input on all 1,000 questions. Accuracy is in \%; change is in percentage points relative to the default setting.}
\label{tab:seed_common_budget}
\begin{tabular}{lrrr}
\toprule
Model & \shortstack{Default native\\video} & \shortstack{1 FPS,\\at most 64 frames} & Change \\
\midrule
Seed-2.0-Pro & 39.60 & 24.00 & $-15.60$ \\
Seed-2.0-Lite & 31.30 & 33.10 & $+1.80$ \\
Seed-2.0-Mini & 27.80 & 26.30 & $-1.50$ \\
\bottomrule
\end{tabular}
\end{table}

Table~\ref{tab:seed_common_budget} shows different responses to the input configuration. Pro drops by 15.60 percentage points, Lite improves by 1.80 points, and Mini decreases by 1.50 points. Under the common frame budget, Lite ranks first, followed by Mini and Pro, reversing Pro's lead under the default native-video interface. These results show that Seed performance depends on the input configuration; its default results reflect both the model and the provider's video-processing pipeline.

\subsection{Critical-event Temporal Localization}
\label{app:event_localization}
% Source consistency note (2026-11-20): values below are preserved from
% MomentVideo_ICLR27_Submit.zip. The localization QA counts sum to 94/300
% (31.33%) for Qwen3.6-27B, 78/300 (26.00%) for Gemma-4-31B, and 63/300
% (21.00%) for InternVL3.5-30B-A3B. The coverage audit reports 30.7%,
% 25.3%, and 20.7%, respectively, while referring to the same subset and
% existing QA predictions. These differences require author verification.

We evaluate temporal localization on the 300 annotated questions containing 609 necessary event intervals. Given the video and question, each model predicts the start and end times of the events supporting its answer. Inputs use 8 FPS candidate sampling with a maximum of 128 frames. Predicted and annotated intervals are matched one-to-one with temporal intersection-over-union (tIoU) at least 0.5, and we report mean localization F1. We also report the fraction of questions for which all necessary events are matched and link these groups to the existing QA predictions for the same questions.

\begin{table}[htbp]
\centering
\small
\caption{Critical-event localization on 300 questions. Interval matching uses tIoU $\geq 0.5$. The last two columns report correct answers / questions and QA accuracy in the all-events-matched group and its complement. F1 and accuracies are in \%.}
\label{tab:event_localization}
\begin{tabular}{lrrrr}
\toprule
Model & \shortstack{Mean\\localization F1} & \shortstack{All events\\matched} & \shortstack{QA: all\\matched} & QA: others \\
\midrule
Qwen3.6-27B & 10.99 & 27/300 (9.00) & 12/27 (44.44) & 82/273 (30.04) \\
Gemma-4-31B & 14.60 & 30/300 (10.00) & 12/30 (40.00) & 66/270 (24.44) \\
InternVL3.5-30B-A3B & 3.68 & 12/300 (4.00) & 4/12 (33.33) & 59/288 (20.49) \\
\bottomrule
\end{tabular}
\end{table}

As shown in Table~\ref{tab:event_localization}, mean localization F1 remains below 15\%, and all necessary events are matched in at most 10\% of questions. An audit of the actual input timestamps finds at least one sampled frame inside approximately 97.0\% of annotated intervals, so interval misses alone cannot account for the low localization scores. QA accuracy is 12.8--15.6 percentage points higher in the all-events-matched group across the three models. Accurate event localization is thus associated with better question answering, while both localization and subsequent use of the evidence remain challenging.

\subsection{Text-only Baselines}
\label{app:text_only}

We evaluate six models on all 1,000 questions with only the question text and, for multiple-choice items, the answer options; no video is provided. Multiple-choice scoring retains the benchmark's requirement that the correct answer be selected under every tested option order. Table~\ref{tab:text_only} compares these results with the video-input results under the main evaluation settings and reports text-only accuracy separately for multiple-choice and open-ended questions.

\begin{table}[htbp]
\centering
\small
\caption{Text-only baselines on all 1,000 questions. Accuracy is in \%; the drop from video to text-only input is in percentage points. MC and OE denote multiple-choice and open-ended questions.}
\label{tab:text_only}
\begin{tabular}{lrrrrr}
\toprule
Model & Video & Text-only & Drop & \shortstack{Text-only\\MC} & \shortstack{Text-only\\OE} \\
\midrule
Seed-2.0-Lite & 31.30 & 14.60 & 16.70 & 26.58 & 10.88 \\
Seed-2.0-Mini & 27.80 & 6.90 & 20.90 & 23.21 & 1.83 \\
Qwen3.6-27B & 22.50 & 8.00 & 14.50 & 25.32 & 2.62 \\
Gemma-4-31B-IT & 19.90 & 3.40 & 16.50 & 10.13 & 1.31 \\
MiMo-V2.5 & 22.10 & 12.70 & 9.40 & 18.14 & 11.01 \\
InternVL3.5-8B & 16.70 & 6.90 & 9.80 & 22.36 & 2.10 \\
\bottomrule
\end{tabular}
\end{table}

Removing the video reduces accuracy by 9.40--20.90 percentage points across all six models, showing that visual input contributes to answering the benchmark questions. Text-only accuracy is higher on multiple-choice items, where the options expose candidate answers and may enable elimination using language priors. Open-ended accuracy is 1.31\%--11.01\%, consistent with the difficulty of recovering specific event counts, actions, and state changes from question text alone.

\subsection{Task-balanced Macro Accuracy}
\label{app:task_macro_average}

The four tasks contain 135 TO, 472 TC, 308 AD, and 85 TR questions. Consequently, TC accounts for 47.2\% of the question-averaged overall score, whereas TR accounts for 8.5\%. We additionally report macro accuracy, which gives each task a weight of 25\%:
\begin{equation}
\mathrm{Acc}_{\mathrm{macro}} = \frac{1}{4}\left(\mathrm{Acc}_{\mathrm{TO}}+\mathrm{Acc}_{\mathrm{TC}}+\mathrm{Acc}_{\mathrm{AD}}+\mathrm{Acc}_{\mathrm{TR}}\right).
\end{equation}
Table~\ref{tab:task_macro_average} reports both aggregates for all 34 models under the main evaluation settings.

\input{tables/task_macro_average}

Seed-2.0-Pro remains first with 42.74\% macro accuracy. Some relative rankings change: DeepSeek-v4.1-Flash scores below Qwen3.5-27B in overall accuracy (22.60\% versus 24.10\%), but above it in macro accuracy (26.41\% versus 24.84\%). Reporting both scores distinguishes performance under the observed task distribution from performance with equal task weights. Future benchmark extensions will include more TR questions requiring integration of brief events and their consequences.

\subsection{Subcategory-level Results}
\label{appendix_result1}

The main paper reports category-level accuracy over seven high-level domains. To provide a more fine-grained view, we further report subcategory-level results over the 25 subcategories in \textbf{\ourmethod}. Table~\ref{tab:subclass_average_accuracy} summarizes the average accuracy of all evaluated models on each subcategory, and Figure~\ref{fig:subcategory_heatmap} visualizes the full model-by-subcategory accuracy matrix.

\input{tables/subclass_result}
\input{figures/heatmap_subclass}

The subcategory-level results show that model performance varies substantially across fine-grained scenarios, even within the same high-level domain. In general, models achieve relatively higher average accuracy on several animal and nature-related subcategories. For example, the average accuracy reaches 45.52\% on \textit{Lightning}, 35.08\% on \textit{Birds}, 34.69\% on \textit{Reptiles}, and 32.58\% on \textit{Mammals}. These subcategories often contain visually salient motion or state changes, such as flying, jumping, lighting variation, or clear object movement, which may be easier to capture when the decisive frames are sampled.

In contrast, several subcategories remain particularly challenging. The average accuracy is low on \textit{Daily Life} (8.59\%), \textit{Music} (10.00\%), \textit{Artifacts} (11.91\%), \textit{Volleyball} (12.57\%), and \textit{Soccer} (13.51\%). These scenarios often involve subtle state changes, small objects, fast camera motion, dense visual layouts, or synthetic visual artifacts, making the momentary event harder to distinguish from background dynamics. Industrial subcategories are also difficult, with 16.33\% on \textit{Manual Assembly} and 13.53\% on \textit{Industrial Machinery}, suggesting that current models struggle with small-region safety-related actions and brief mechanical movements.

The results also reveal large variation within the same domain. For example, within GUI videos, \textit{Operating System} achieves 31.29\% average accuracy, while \textit{Website} and \textit{Application} remain much lower at 17.32\% and 15.11\%, respectively. Similarly, within human activities, \textit{Non-ball Sports} reaches 24.78\%, whereas \textit{Daily Life} is only 8.59\%. This indicates that coarse category-level accuracy can hide important differences among fine-grained scenarios. We note that several subcategories contain relatively few samples, such as \textit{Fish}, \textit{Amphibians}, \textit{XR}, and \textit{Invertebrates}; their average accuracies should therefore be interpreted as indicative rather than conclusive.

Overall, the subcategory-level analysis complements the main results by showing that momentary visual event understanding is highly scenario-dependent and that fine-grained evaluation is necessary for revealing model weaknesses hidden by broad domain-level metrics.

\subsection{Duration-wise Results for All Models}
\label{appendix_result2}
In addition to the duration analysis of three representative models in the main paper, we provide model-level duration-wise results for all evaluated models. Figure~\ref{fig:duration_all_models} shows the accuracy of each model across six video-duration buckets: $\leq$5s, 5--10s, 10--20s, 20--60s, 1--3min, and $\geq$3min. Each subplot corresponds to one model, and each bar represents the accuracy within a specific duration bucket.

Overall, the model-level results are consistent with the general trend observed in the main paper. Most models achieve relatively better performance on short-to-medium videos, especially in the 5--10s and 10--20s buckets, while performance generally drops on longer videos. The reported average accuracy reaches 21.51\% on 5--10s videos and 21.99\% on 10--20s videos, compared with 18.47\% on videos shorter than 5 seconds. Surrounding context may contribute to this pattern, but the duration groups also differ in task and content composition. Performance decreases as videos become longer. The average accuracy drops to 16.95\% on 20--60s videos, 11.97\% on 1--3min videos, and 4.18\% on videos longer than 3 minutes. This pattern is consistent with greater difficulty locating relevant evidence in longer videos. However, these comparisons do not directly measure temporal localization or internal aggregation, and do not isolate duration from differences in sample composition. We also observe that the degradation pattern varies across models. Some models remain relatively stable across short- and medium-duration buckets, while others are more sensitive to duration changes. However, the longest-duration bucket remains consistently difficult for most models. Since the two longest buckets contain relatively few samples, their results should be interpreted as indicative trends rather than definitive conclusions. Overall, these findings establish an association between video duration and answer accuracy on this benchmark, without identifying the internal mechanism responsible for the performance differences.

\input{figures/duration_subplot}

\section{Partial Model Test Results}
\label{appendix_partial_results}

\begin{figure*}[t]
    \centering
    \includegraphics[width=\textwidth,height=0.93\textheight,keepaspectratio]{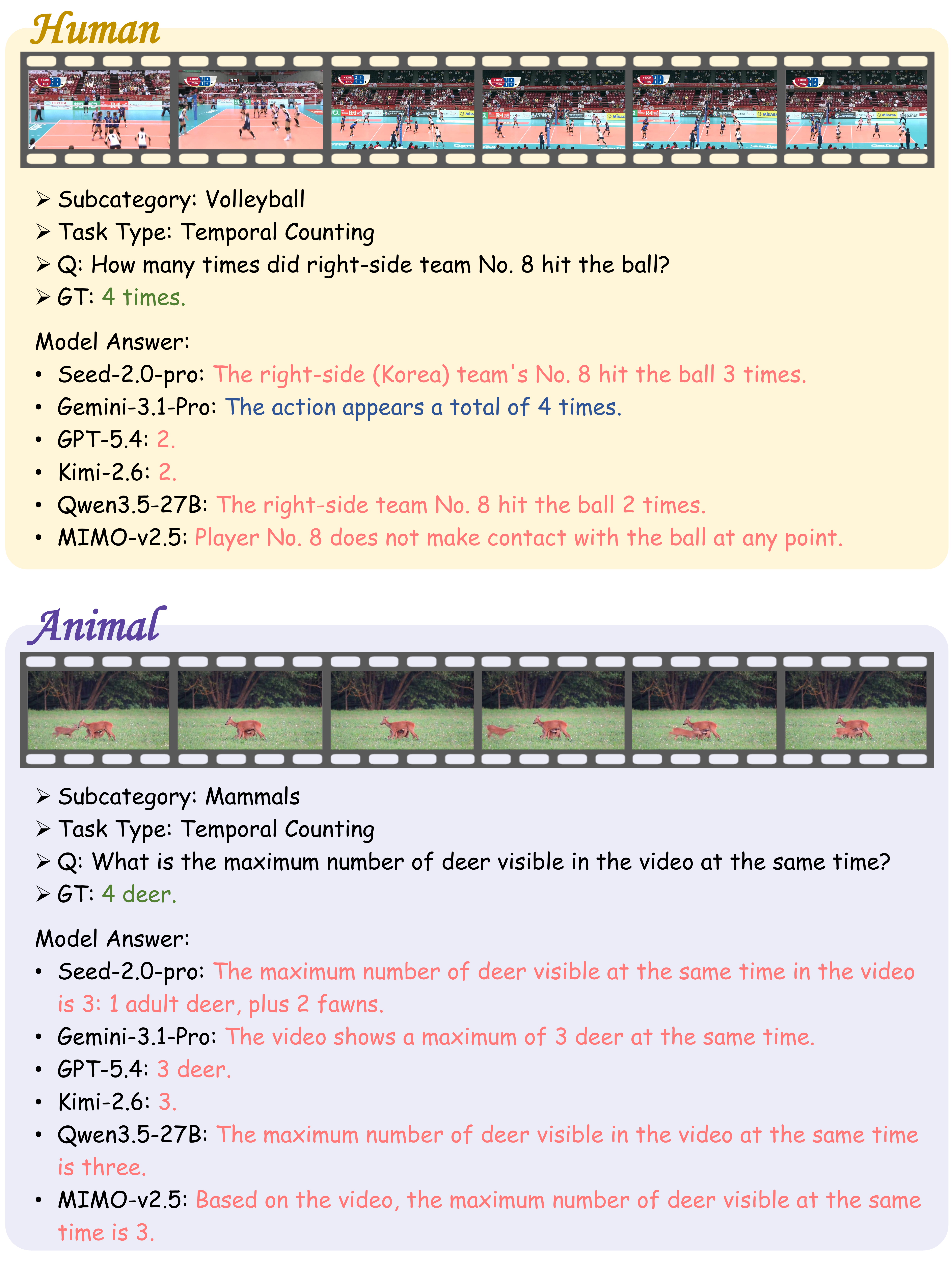}
    \caption{
    Partial model test results on \textbf{\ourmethod} (Part I).
    }
    \label{fig:partial_results}
\end{figure*}

\begin{figure*}[t]
    \centering
    \includegraphics[width=\textwidth,height=0.93\textheight,keepaspectratio]{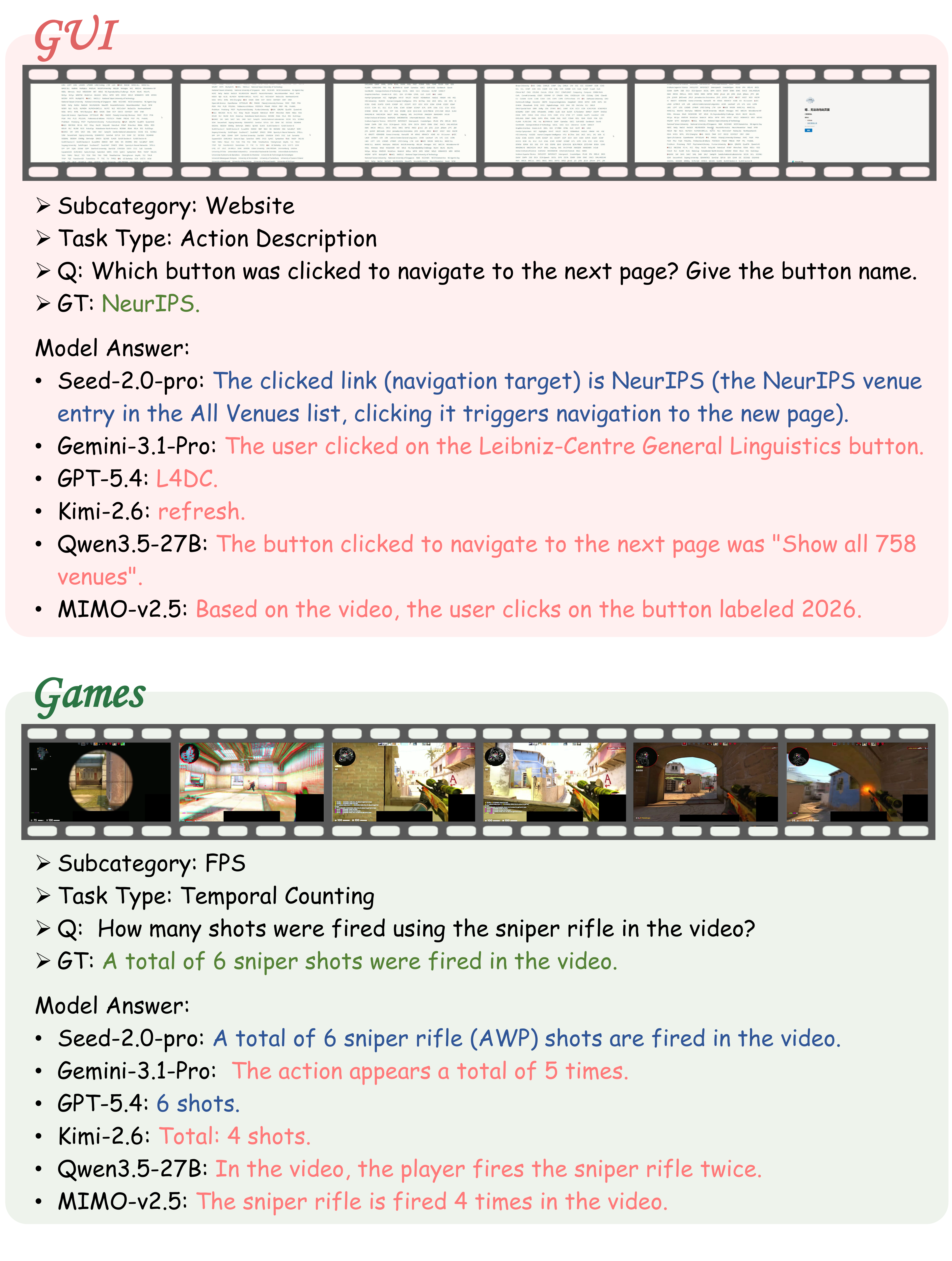}
    \caption{
    Partial model test results on \textbf{\ourmethod} (Part II).
    }
    \label{fig:partial_results2}
\end{figure*}

\begin{figure*}[t]
    \centering
    \includegraphics[width=\textwidth,height=0.93\textheight,keepaspectratio]{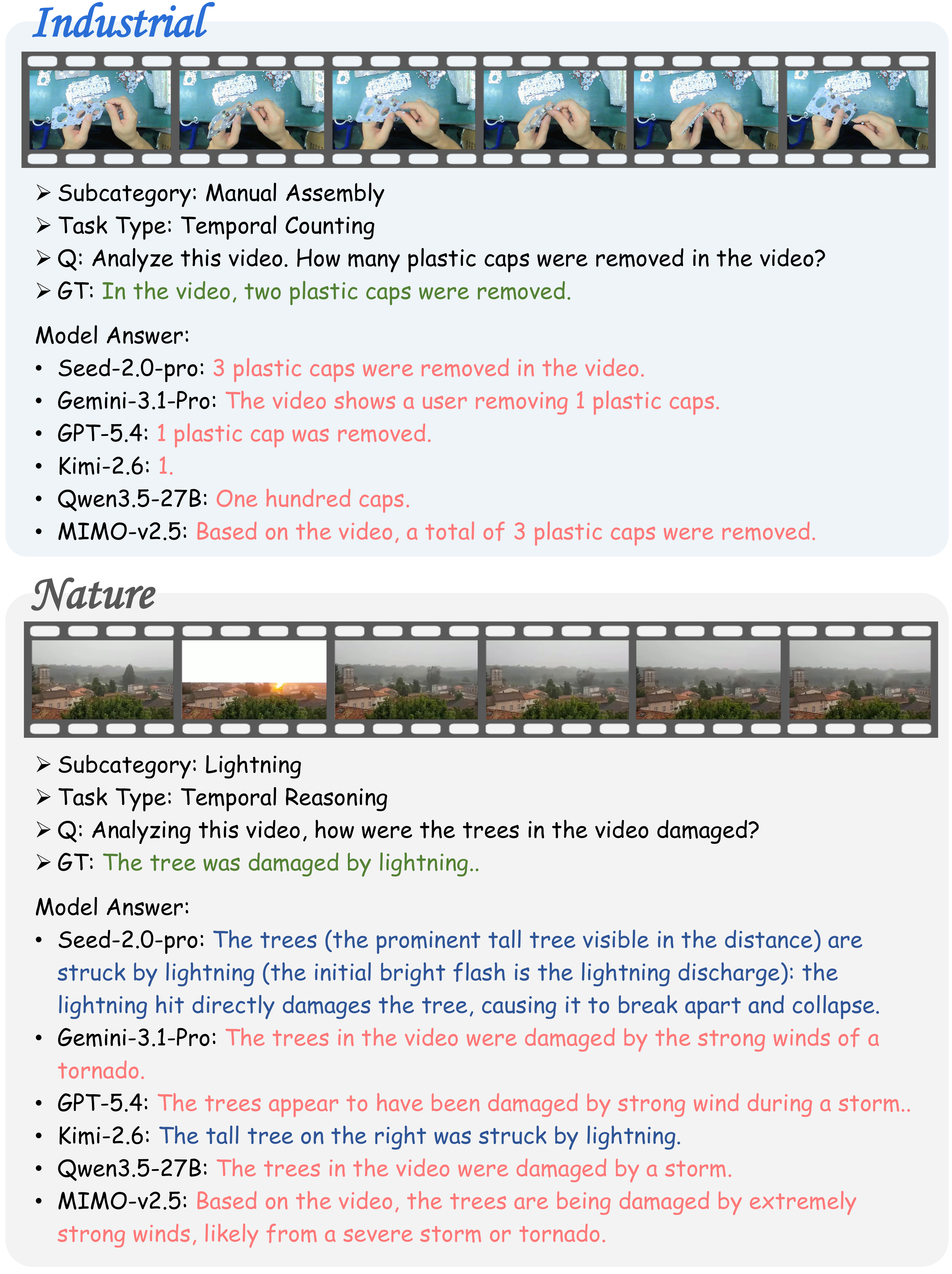} %
    \caption{
    Partial model test results on \textbf{\ourmethod} (Part III).
    }
    \label{fig:partial_results3}
\end{figure*}

\begin{figure*}[t]
    \centering
    \includegraphics[width=\textwidth]{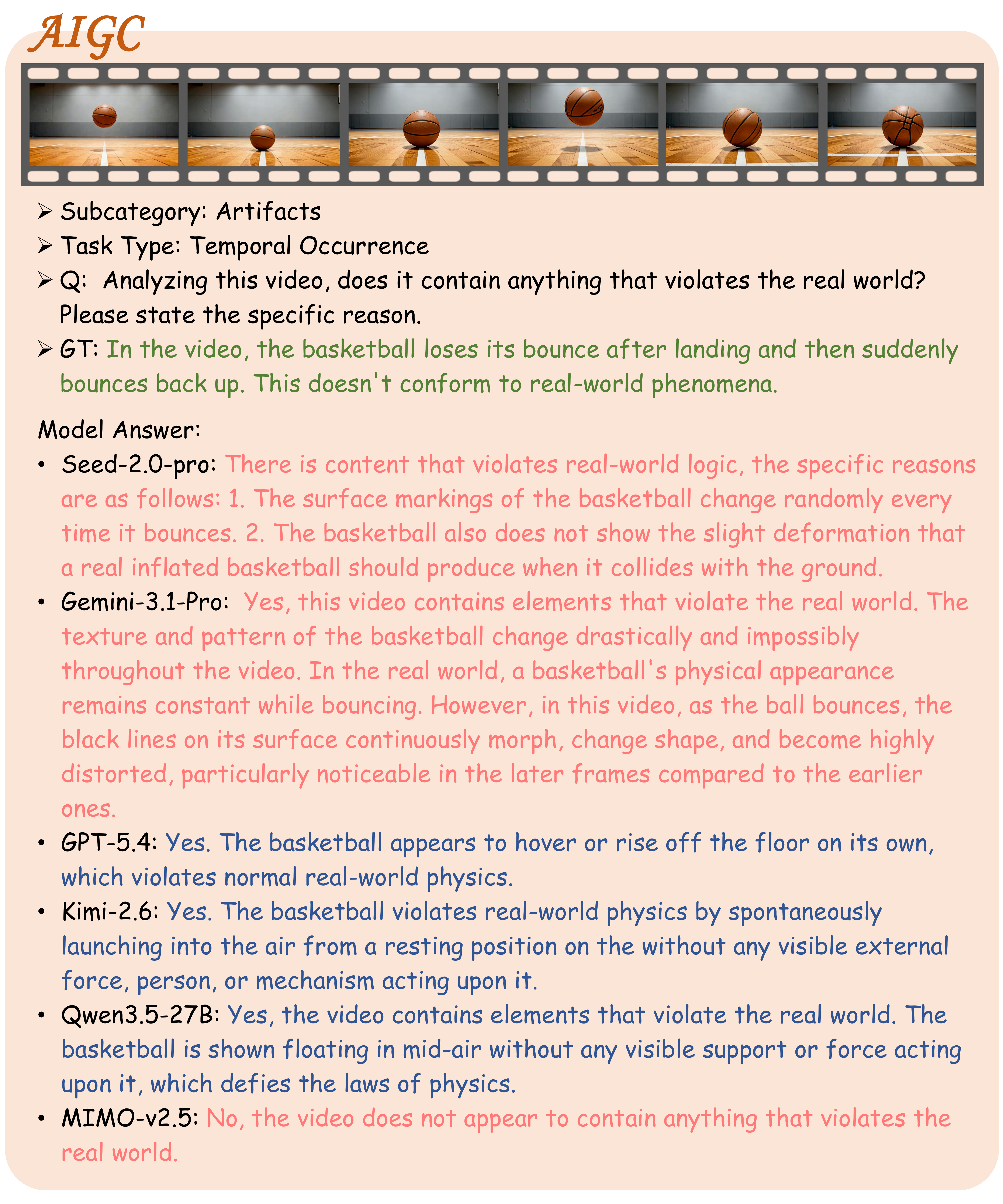} %
    \caption{
    Partial model test results on \textbf{\ourmethod} (Part IV).
    }
    \label{fig:partial_results4}
\end{figure*}

We provide partial model test results in Figures~\ref{fig:partial_results}, ~\ref{fig:partial_results2}, ~\ref{fig:partial_results3}, and ~\ref{fig:partial_results4}. These examples show representative outputs of different models on \ourmethod.

\input{tables/prompt_open}
\section{LLM-as-Judge Prompt}
\label{appendix_prompt}
{\ourmethod} contains both multiple-choice and open-ended questions, and we adopt different evaluation strategies for these two formats. All multiple-choice questions have exactly one correct option. A question with $K$ options is evaluated under $K$ shuffled option orders (e.g., four evaluations for four options). The model response is normalized and matched to the ground-truth option under each order. A sample is counted as correct only if the model selects the semantically correct option in every evaluation.

For open-ended questions, exact string matching is often too strict, since semantically equivalent answers may appear in different surface forms. Therefore, we use an LLM-as-Judge protocol only for open-ended questions. The judge compares the model prediction with the canonical reference answer and determines whether they are semantically consistent. The judge does not see the video, does not re-answer the question, and is not allowed to introduce external knowledge. It is used solely for answer-equivalence checking. All judge calls are performed with temperature 0, and the judge is required to output a structured JSON object. As shown in Figure~\ref{fig:open_judge_prompt}, the judge only evaluates semantic consistency between the model prediction and the reference answer.

%% file: tables/data_source.tex
\begin{table*}[t]
\centering
\footnotesize
\setlength{\tabcolsep}{4pt}
\renewcommand{\arraystretch}{1.08}
\caption{
Data sources of the fine-grained subcategories in \textbf{\ourmethod}. 
}
\label{tab:data_sources}
\begin{tabularx}{\textwidth}{p{0.11\textwidth}p{0.25\textwidth}X}
\toprule
\textbf{Domain} & \textbf{Subcategory} & \textbf{Source} \\
\midrule
AIGC 
& Artifacts 
& Publicly available subset from RISEBench~\citep{risebench}. \\

\midrule
GUI 
& XR 
& Publicly available subset from GUI-World~\citep{gui_xr}. \\

GUI 
& Website, Application
& Publicly available subset from GUI-World~\citep{gui_xr} and self-collected screen recordings. \\

GUI 
& Operating System 
& Self-collected screen recordings. \\

\midrule
Nature 
& Lightning 
& Publicly available video dataset subset~\citep{light} and YouTube clips. \\

Nature 
& Other Nature Phenomena 
& YouTube clips. \\

\midrule
Industry 
& Manual Assembly
& Publicly available subset from Egocentric-10K~\citep{workerdata}. \\

Industry 
& Industrial Machinery
& YouTube clips. \\

\midrule
Games 
& FPS 
& YouTube gameplay clips from CS2 and Valorant. \\

Games 
& MOBA 
& YouTube gameplay clips from League of Legends. \\

Games 
& Combat 
& YouTube gameplay clips from Street Fighter 6, Tekken 8, and Guilty Gear. \\

Games 
& Music 
& YouTube gameplay clips from osu!, Phigros, and Muse Dash. \\

\midrule
Human 
& Basketball, Soccer, Volleyball 
& Publicly available subsets from MultiSports~\citep{multisports} and SVW~\citep{SVW}. \\

Human 
& Other Ball Sports, Non-ball Sports, Daily Life
& Publicly available subsets from MultiSports~\citep{multisports}, SVW~\citep{SVW}, the shuttlecock trajectory dataset\citep{badminton} and UCF101~\citep{UCF101}. \\

\midrule
Animal 
& Birds, Mammals, Reptiles, Amphibians, Fish, Invertebrates
& Publicly available subset from Animal Kingdom~\citep{ani_kingdom}. \\

\bottomrule
\end{tabularx}
\end{table*}

%% file: tables/appendix_dataStatic.tex
\begin{table}[t]
\centering
\footnotesize
\setlength{\tabcolsep}{4pt}
\renewcommand{\arraystretch}{0.98}
\caption{
\textbf{\ourmethod} domain and subcategory stats (one QA per video).
}
\label{tab:subcategory_statistics}
\begin{tabular}{llc}
\toprule
\textbf{Domain (\#)} & \textbf{Subcategory} & \textbf{\#Videos / \#QA} \\
\midrule
\multirow{1}{*}{AIGC (79)}
& Artifacts & 79 \\
\midrule
\multirow{4}{*}{GUI (218)}
& XR & 8 \\
& Website & 97 \\
& Application & 91 \\
& Operating System & 22 \\
\midrule
\multirow{2}{*}{Nature (72)}
& Lightning & 49 \\
& Other Natural Phenomena & 23 \\
\midrule
\multirow{2}{*}{Industry (121)}
& Manual Assembly & 96 \\
& Industrial Machinery & 25 \\
\midrule
\multirow{4}{*}{Games (160)}
& FPS & 60 \\
& MOBA & 50 \\
& Combat & 30 \\
& Music & 20 \\
\midrule
\multirow{6}{*}{Human (250)}
& Basketball & 57 \\
& Soccer & 38 \\
& Volleyball & 40 \\
& Other Ball Sports & 37 \\
& Non-ball Sports & 40 \\
& Daily Life & 38 \\
\midrule
\multirow{6}{*}{Animal (100)}
& Birds & 42 \\
& Mammals & 13 \\
& Reptiles & 29 \\
& Amphibians & 6 \\
& Fish & 2 \\
& Invertebrates & 8 \\
\midrule
\textbf{Total} & -- & \textbf{1,000} \\
\bottomrule
\end{tabular}
\end{table}

\begin{table}[t]
\centering
\small
\setlength{\tabcolsep}{6pt}
\renewcommand{\arraystretch}{1.05}
\caption{
Task-type statistics of \textbf{\ourmethod}.
}
\label{tab:task_statistics}
\begin{tabular}{lcc}
\toprule
\textbf{Task Type} & \textbf{Abbr.} & \textbf{\#Questions} \\
\midrule
Temporal Occurrence & TO & 135 \\
Temporal Counting & TC & 472 \\
Action Description & AD & 308 \\
Temporal Reasoning & TR & 85 \\
\midrule
\textbf{Total} & -- & \textbf{1,000} \\
\bottomrule
\end{tabular}
\end{table}

\begin{table}[t]
\centering
\small
\setlength{\tabcolsep}{6pt}
\renewcommand{\arraystretch}{1.05}
\caption{
Answer-format statistics of \textbf{\ourmethod}.
}
\label{tab:answer_format_statistics}
\begin{tabular}{lc}
\toprule
\textbf{Answer Format} & \textbf{\#Questions} \\
\midrule
Open-ended & 764 \\
Multiple-choice & 236 \\
\midrule
\textbf{Total} & \textbf{1,000} \\
\bottomrule
\end{tabular}
\end{table}

%% file: tables/result_8fps.tex
\begin{table*}[t]
\centering
\small
\caption{
Main results on \textbf{\ourmethod}. 
The \textbf{bold} and \underline{underlined} numbers indicate the best and second-best performance within each group. 
Gray rows indicate additional 8 FPS results. Frame-sequence inputs use at most 128 frames (GPT-5.4: 50); native video models use their supported interfaces.
}

\setlength{\tabcolsep}{3.5pt}
\resizebox{0.99\textwidth}{!}{
\begin{tabular}{lccccccccccccc}
\toprule
\multirow{2}{*}{\textbf{Model}} 
& \textbf{Input}
& \multicolumn{4}{c}{\textbf{Task Type(\%)}}
& \multicolumn{7}{c}{\textbf{Video Category(\%)}}
& \multirow{2}{*}{\textbf{Overall(\%)}} \\
\cmidrule(lr){3-6} \cmidrule(lr){7-13}
& \textbf{Settings}
& \textbf{TO} & \textbf{TC} & \textbf{AD} & \textbf{TR} 
& \textbf{AIGC} & \textbf{GUI} & \textbf{Nature} & \textbf{Industry} & \textbf{Games} & \textbf{Human} & \textbf{Animal} & \\
\midrule
\rowcolor{gray!15} \multicolumn{14}{c}{\textbf{Proprietary Models}} \\
\midrule

Seed-2.0-Pro~\cite{seed}
& default & \textbf{50.37} & \textbf{31.14} & \underline{47.08} & \underline{42.35} & \textbf{45.57} & 33.94 & \textbf{69.44} & \textbf{41.32} & \textbf{39.37} & \underline{27.20} & 55.00 & \textbf{39.6} \\

\rowcolor{gray!10}
Gemini-3-Flash~\cite{gemini3}
& 8fps & 40.00 & \underline{27.97} & \textbf{49.03} & 38.82 & \underline{37.97} & \textbf{37.61} & 56.94 & 26.45 & \underline{35.63} & \textbf{28.00} & \underline{58.00} & \underline{37.0} \\

\rowcolor{gray!10}
Gemini-3.1-Pro~\cite{gemini3}
& 8fps & 40.00 & 27.75 & 45.45 & \textbf{44.71} & 35.44 & \underline{34.86} & \underline{58.33} & \underline{38.02} & 35.00 & 23.60 & 56.00 & 36.3 \\

Seed-2.0-Lite~\cite{seed}
& default & 34.81 & 25.64 & 36.04 & 40.00 & 22.78 & 27.98 & 51.39 & 30.58 & 25.62 & 26.80 & 52.00 & 31.3 \\

\rowcolor{gray!10}
Gemini-3.1-Flash-Lite~\cite{gemini3}
& 8fps & 34.81 & 24.79 & 37.34 & 38.82 & 27.85 & 27.98 & 55.56 & 29.75 & 25.00 & 21.20 & \textbf{60.00} & 31.2 \\

\rowcolor{gray!10}
GPT-5.4~\cite{GPT5}
& 8fps & 32.59 & 21.19 & 40.26 & 35.29 & 22.78 & 34.40 & 41.67 & 19.83 & 29.38 & 20.00 & 54.00 & 29.8 \\

Seed-2.0-Mini~\cite{seed}
& default & 32.59 & 22.88 & 33.77 & 25.88 & 24.05 & 22.94 & \underline{58.33} & 29.75 & 22.50 & 20.80 & 43.00 & 27.8 \\

Gemini-3.1-Pro~\cite{gemini3}
& 1fps & 32.59 & 19.70 & 33.44 & 34.12 & 27.85 & 27.52 & 44.44 & 18.18 & 24.37 & 19.20 & 46.00 & 26.9 \\

Gemini-3-Flash~\cite{gemini3}
& 1fps & \underline{41.48} & 18.43 & 31.82 & 32.94 & \textbf{45.57} & 20.64 & 43.06 & 15.70 & 20.62 & 23.20 & 47.00 & 26.9 \\

Gemini-3.1-Flash-Lite~\cite{gemini3}
& 1fps & 34.07 & 18.64 & 30.52 & 27.06 & 27.85 & 22.02 & 41.67 & 22.31 & 20.00 & 14.00 & 57.00 & 25.1 \\

GPT-5.4~\cite{GPT5}
& 1fps & 21.48 & 12.08 & 30.19 & 29.41 & 6.33 & 23.85 & 40.28 & 4.96 & 18.75 & 18.00 & 37.00 & 20.4 \\

\midrule
\rowcolor{gray!15} \multicolumn{14}{c}{\textbf{Open-source Models}} \\
\midrule

\rowcolor{gray!10}
DeepSeek-v4.1-Flash~\cite{ds41}
& 8fps & \underline{36.30} & 19.49 & \textbf{38.96} & \textbf{34.12} & 26.58 & \textbf{40.83} & 51.39 & 17.36 & 19.38 & \textbf{20.80} & 39.00 & \textbf{29.0} \\

\rowcolor{gray!10}
Qwen3.5-122B-A10B~\cite{qwen3_5}
& 8fps & \textbf{39.26} & 19.49 & 34.42 & 27.06 & \textbf{39.24} & 23.39 & 38.89 & \textbf{23.14} & 21.25 & \underline{20.40} & 51.00 & \underline{27.4} \\

\rowcolor{gray!10}
Qwen3.6-27B~\cite{qwen3.6-27b}
& 8fps & 28.89 & \underline{21.40} & 34.74 & 30.59 & 24.05 & 26.15 & 45.83 & \textbf{23.14} & \textbf{23.75} & 19.20 & 50.00 & 27.3 \\

\rowcolor{gray!10}
Qwen3.5-397B-A17B~\cite{qwen3_5}
& 8fps & 29.63 & 17.16 & \underline{36.69} & 25.88 & 18.99 & 24.31 & 50.00 & 13.22 & 20.00 & 19.60 & \textbf{55.00} & 25.6 \\

\rowcolor{gray!10}
Qwen3-VL-235B-A22B~\cite{qwen3_vl}
& 8fps & 16.30 & \textbf{22.67} & 33.12 & 28.24 & 5.06 & 26.61 & 50.00 & 21.49 & \underline{22.50} & 19.60 & 46.00 & 25.5 \\

\rowcolor{gray!10}
Qwen3.5-27B~\cite{qwen3_5}
& 8fps & 22.22 & 19.92 & 36.36 & 20.00 & 12.66 & 24.31 & \underline{52.78} & 20.66 & \underline{22.50} & 15.20 & 53.00 & 25.3 \\

Kimi-2.6~\cite{kimi2.6}
& 1fps & 25.93 & 15.89 & \underline{36.69} & 30.59 & 15.19 & \underline{37.16} & 41.67 & 12.40 & 16.25 & \underline{20.40} & 34.00 & 24.9 \\

\rowcolor{gray!10}
Qwen3.6-35B-A3B~\cite{qwen3.6-35b-a3b}
& 8fps & 28.15 & 19.49 & 32.47 & 21.18 & 21.52 & 22.48 & 40.28 & 21.49 & 20.62 & 18.40 & 48.00 & 24.8 \\

\rowcolor{gray!10}
Gemma-4-31B~\cite{gemma4}
& 8fps & 32.59 & 17.16 & 31.17 & 27.06 & 26.58 & 16.97 & 47.22 & 17.36 & \textbf{23.75} & 15.60 & \underline{54.00} & 24.4 \\

Qwen3.5-27B~\cite{qwen3_5}
& 1fps & 20.74 & 19.07 & 32.47 & 27.06 & 11.39 & 21.56 & 41.67 & 18.18 & 18.75 & 20.00 & 53.00 & 24.1 \\

\rowcolor{gray!10}
Qwen3.5-35B-A3B~\cite{qwen3_5}
& 8fps & 19.26 & 17.58 & 36.04 & 21.18 & 8.86 & 23.85 & \textbf{55.56} & 17.36 & 16.25 & 18.00 & 47.00 & 23.8 \\

\rowcolor{gray!10}
Qwen3.5-9B~\cite{qwen3_5}
& 8fps & 34.81 & 15.89 & 31.17 & 20.00 & \underline{35.44} & 22.02 & 50.00 & 12.40 & 13.75 & 15.20 & 48.00 & 23.5 \\

Qwen3.5-397B-A17B~\cite{qwen3_5}
& 1fps & 27.41 & 14.83 & 32.79 & 23.53 & 16.46 & 23.39 & 47.22 & 18.18 & 17.50 & 14.80 & 43.00 & 22.8 \\

DeepSeek-v4.1-Flash~\cite{ds41}
& 1fps & 30.37 & 14.62 & 28.90 & \underline{31.76} & 27.85 & 32.57 & 41.67 & 9.92 & 12.50 & 17.20 & 28.00 & 22.6 \\

Qwen3.6-27B~\cite{qwen3.6-27b}
& 1fps & 19.26 & 17.16 & 31.82 & 23.53 & 10.13 & 26.61 & 41.67 & 19.01 & 16.25 & 14.40 & 44.00 & 22.5 \\

MIMO-v2.5~\cite{mimov25}
& default & 18.52 & 17.16 & 30.84 & 23.53 & 2.53 & 24.31 & \underline{52.78} & 19.01 & 19.38 & 14.80 & 37.00 & 22.1 \\

Qwen3.5-122B-A10B~\cite{qwen3_5}
& 1fps & 25.19 & 14.19 & 27.60 & 23.53 & 18.99 & 22.02 & 40.28 & 17.36 & 13.75 & 14.40 & 35.00 & 20.6 \\

Qwen3.6-35B-A3B~\cite{qwen3.6-35b-a3b}
& 1fps & 17.04 & 14.41 & 29.22 & 24.71 & 3.80 & 18.35 & 41.67 & \underline{22.31} & 15.62 & 15.20 & 39.00 & 20.2 \\

Qwen3.5-35B-A3B~\cite{qwen3_5}
& 1fps & 20.00 & 16.10 & 26.30 & 18.82 & 8.86 & 19.27 & 45.83 & 19.83 & 13.75 & 12.80 & 40.00 & 20.0 \\

Gemma-4-31B~\cite{gemma4}
& 1fps & 30.37 & 13.56 & 24.35 & 22.35 & 27.85 & 11.93 & 41.67 & 17.36 & 15.62 & 15.60 & 36.00 & 19.9 \\

\rowcolor{gray!10}
Qwen3-VL-30B-A3B~\cite{qwen3_vl}
& 8fps & 15.56 & 15.47 & 26.30 & 22.35 & 7.59 & 16.51 & 48.61 & 15.70 & 18.12 & 13.20 & 36.00 & 19.4 \\

\rowcolor{gray!10}
InternVL3.5-241B-A28B~\cite{internvl35}
& 8fps & 18.52 & 14.41 & 25.65 & 20.00 & 6.33 & 12.84 & 45.83 & 7.44 & 20.60 & 15.20 & 43.00 & 18.9 \\

\rowcolor{gray!10}
Qwen3.5-4B~\cite{qwen3_5}
& 8fps & 17.04 & 17.58 & 23.05 & 11.76 & 7.59 & 15.60 & 44.44 & 14.05 & 13.13 & 14.80 & 40.00 & 18.7 \\

Qwen3.5-9B~\cite{qwen3_5}
& 1fps & 18.52 & 13.35 & 26.62 & 18.82 & 10.13 & 17.43 & 40.28 & 12.40 & 11.87 & 14.80 & 40.00 & 18.6 \\

\rowcolor{gray!10}
Qwen3-VL-8B~\cite{qwen3_vl}
& 8fps & 20.00 & 15.04 & 22.73 & 18.82 & 16.46 & 16.06 & 41.67 & 11.57 & 11.25 & 13.20 & 41.00 & 18.4 \\

\rowcolor{gray!10}
InternVL3.5-30B-A3B~\cite{internvl35}
& 8fps & 17.78 & 14.19 & 23.70 & 16.47 & 6.33 & 12.84 & 44.44 & 19.10 & 16.88 & 12.80 & 31.00 & 17.8 \\

\rowcolor{gray!10}
InternVL3.5-8B~\cite{internvl35}
& 8fps & 14.81 & 16.10 & 20.78 & 16.47 & 2.53 & 10.55 & 41.67 & 16.53 & 16.88 & 13.60 & 38.00 & 17.4 \\

Qwen3.5-4B~\cite{qwen3_5}
& 1fps & 19.26 & 13.77 & 21.43 & 17.65 & 8.86 & 16.97 & 38.89 & 13.22 & 15.00 & 11.20 & 32.00 & 17.2 \\

\rowcolor{gray!10}
InternVL3.5-4B~\cite{internvl35}
& 8fps & 15.56 & 15.89 & 21.43 & 10.59 & 5.06 & 11.47 & 31.94 & 14.88 & 15.62 & 13.60 & 42.00 & 17.1 \\

Qwen3-VL-235B-A22B~\cite{qwen3_vl}
& 1fps & 15.56 & 12.08 & 23.38 & 23.53 & 5.06 & 16.97 & 38.89 & 10.74 & 10.62 & 14.00 & 36.00 & 17.0 \\

InternVL3.5-30B-A3B~\cite{internvl35}
& 1fps & 12.59 & 16.10 & 21.75 & 11.76 & 3.80 & 10.55 & 38.89 & 15.70 & 15.00 & 18.40 & 27.00 & 17.0 \\

Qwen3-VL-30B-A3B~\cite{qwen3_vl}
& 1fps & 18.52 & 13.14 & 20.45 & 20.00 & 8.86 & 14.22 & 43.06 & 14.05 & 15.62 & 9.60 & 32.00 & 16.7 \\

InternVL3.5-8B~\cite{internvl35}
& 1fps & 15.56 & 15.04 & 20.13 & 15.29 & 3.80 & 11.93 & 38.89 & 13.22 & 13.75 & 14.00 & 37.00 & 16.7 \\

LLaVA-Video-72B~\cite{llava}
& 1fps & 13.33 & 15.04 & 18.83 & 17.65 & 0.00 & 8.26 & 34.72 & 19.83 & 16.25 & 15.20 & 31.00 & 16.2 \\

InternVL3.5-241B-A28B~\cite{internvl35}
& 1fps & 19.26 & 11.02 & 20.45 & 18.82 & 1.27 & 12.39 & 40.28 & 9.09 & 11.87 & 13.60 & 36.00 & 15.7 \\

Keye-VL-1.5-8B~\cite{keye}
& 1fps & 14.81 & 12.92 & 20.45 & 14.12 & 0.00 & 10.09 & 29.17 & 17.36 & 14.37 & 14.40 & 33.00 & 15.6 \\

\rowcolor{gray!10}
GLM-4.6V~\cite{glm45v}
& 8fps & 14.81 & 9.32 & 21.75 & 20.00 & 2.53 & 11.47 & 36.11 & 9.92 & 10.62 & 13.60 & 34.00 & 14.80 \\

InternVL3.5-4B~\cite{internvl35}
& 1fps & 14.81 & 13.14 & 17.86 & 7.06 & 3.80 & 10.55 & 29.17 & 14.05 & 13.75 & 10.80 & 30.00 & 14.3 \\

\rowcolor{gray!10}
Qwen3-VL-4B~\cite{qwen3_vl}
& 8fps & 15.56 & 10.38 & 19.48 & 14.12 & 3.80 & 11.47 & 44.44 & 9.92 & 9.38 & 8.80 & 33.00 & 14.2 \\

GLM-4.6V~\cite{glm45v}
& 1fps & 15.56 & 8.69 & 19.81 & 21.18 & 5.06 & 10.55 & 31.94 & 7.44 & 11.25 & 12.40 & 33.00 & 14.1 \\

Qwen3-VL-8B~\cite{qwen3_vl}
& 1fps & 15.56 & 10.59 & 17.86 & 16.47 & 8.86 & 13.30 & 41.67 & 9.92 & 6.88 & 8.80 & 29.00 & 14.0 \\

Qwen3-VL-4B~\cite{qwen3_vl}
& 1fps & 14.81 & 8.69 & 19.48 & 21.18 & 1.27 & 14.22 & 41.67 & 7.44 & 9.38 & 9.20 & 30.00 & 13.9 \\

\rowcolor{gray!10}
Keye-VL-1.5-8B~\cite{keye}
& 8fps & 15.56 & 11.44 & 17.86 & 9.41 & 2.53 & 7.80 & 31.94 & 12.40 & 10.62 & 14.80 & 27.00 & 13.8 \\

\rowcolor{gray!10}
LLaVA-Video-7B~\cite{llava}
& 8fps & 11.11 & 12.92 & 15.26 & 10.59 & 1.27 & 8.26 & 31.94 & 12.40 & 16.88 & 10.80 & 21.00 & 13.2 \\

LLaVA-Video-7B~\cite{llava}
& 1fps & 11.11 & 13.14 & 13.96 & 7.06 & 0.00 & 8.72 & 23.61 & 8.26 & 15.62 & 12.80 & 23.00 & 12.6 \\

VideoLLaMA3-7B~\cite{videollama3}
& 1fps & 7.52 & 13.69 & 11.72 & 8.33 & 0.00 & 10.44 & 25.35 & 4.96 & 10.14 & 15.64 & 15.00 & 11.8 \\

\rowcolor{gray!10}
GLM-4.6V-Flash~\cite{glm45v}
& 8fps & 12.59 & 6.99 & 15.26 & 18.82 & 0.00 & 8.62 & 45.83 & 14.05 & 10.00 & 6.00 & 14.00 & 11.3 \\

GLM-4.6V-Flash~\cite{glm45v}
& 1fps & 11.11 & 7.20 & 14.29 & 20.00 & 1.27 & 8.72 & 43.06 & 13.22 & 10.00 & 6.00 & 12.00 & 11.0 \\

VITA-1.5~\cite{vita1.5}
& 1fps & 19.26 & 8.49 & 13.42 & 10.59 & 0.00 & 5.31 & 25.00 & 8.26 & 11.25 & 8.80 & 26.00 & 10.6 \\

\rowcolor{gray!10}
VideoLLaMA3-7B~\cite{videollama3}
& 8fps & 8.89 & 11.44 & 8.12 & 5.88 & 0.00 & 4.59 & 22.22 & 7.44 & 11.87 & 12.40 & 11.00 & 9.6 \\

\bottomrule
\end{tabular}
}
\label{tab:8fps_result}
\end{table*}

%% file: tables/task_macro_average.tex
\begin{table}[p]
\centering
\small
\setlength{\tabcolsep}{4pt}
\caption{Task-balanced macro accuracy under the main evaluation settings. Overall averages correctness over all 1,000 questions; Macro averages the four task accuracies with equal weight. All scores are in \%. Rows follow macro rank, with shared ranks for equal displayed scores.}
\label{tab:task_macro_average}
\begin{tabular}{rlrrrrrr}
\toprule
Rank & Model & TO & TC & AD & TR & Overall & Macro \\
\midrule
1 & Seed-2.0-Pro & 50.37 & 31.14 & 47.08 & 42.35 & 39.60 & 42.74 \\
2 & Seed-2.0-Lite & 34.81 & 25.64 & 36.04 & 40.00 & 31.30 & 34.12 \\
3 & Gemini-3-Flash & 41.48 & 18.43 & 31.82 & 32.94 & 26.90 & 31.17 \\
4 & Gemini-3.1-Pro & 32.59 & 19.70 & 33.44 & 34.12 & 26.90 & 29.96 \\
5 & Seed-2.0-Mini & 32.59 & 22.88 & 33.77 & 25.88 & 27.80 & 28.78 \\
6 & Gemini-3.1-Flash-Lite & 34.07 & 18.64 & 30.52 & 27.06 & 25.10 & 27.57 \\
7 & Kimi-2.6 & 25.93 & 15.89 & 36.69 & 30.59 & 24.90 & 27.28 \\
8 & DeepSeek-v4.1-Flash & 30.37 & 14.62 & 28.90 & 31.76 & 22.60 & 26.41 \\
9 & Qwen3.5-27B & 20.74 & 19.07 & 32.47 & 27.06 & 24.10 & 24.84 \\
10 & Qwen3.5-397B-A17B & 27.41 & 14.83 & 32.79 & 23.53 & 22.80 & 24.64 \\
11 & GPT-5.4 & 21.48 & 12.08 & 30.19 & 29.41 & 20.40 & 23.29 \\
12 & Qwen3.6-27B & 19.26 & 17.16 & 31.82 & 23.53 & 22.50 & 22.94 \\
13 & Gemma-4-31B & 30.37 & 13.56 & 24.35 & 22.35 & 19.90 & 22.66 \\
14 & Qwen3.5-122B-A10B & 25.19 & 14.19 & 27.60 & 23.53 & 20.60 & 22.63 \\
15 & MIMO-v2.5 & 18.52 & 17.16 & 30.84 & 23.53 & 22.10 & 22.51 \\
16 & Qwen3.6-35B-A3B & 17.04 & 14.41 & 29.22 & 24.71 & 20.20 & 21.35 \\
17 & Qwen3.5-35B-A3B & 20.00 & 16.10 & 26.30 & 18.82 & 20.00 & 20.31 \\
18 & Qwen3.5-9B & 18.52 & 13.35 & 26.62 & 18.82 & 18.60 & 19.33 \\
19 & Qwen3-VL-235B-A22B & 15.56 & 12.08 & 23.38 & 23.53 & 17.00 & 18.64 \\
20 & Qwen3.5-4B & 19.26 & 13.77 & 21.43 & 17.65 & 17.20 & 18.03 \\
20 & Qwen3-VL-30B-A3B & 18.52 & 13.14 & 20.45 & 20.00 & 16.70 & 18.03 \\
22 & InternVL3.5-241B-A28B & 19.26 & 11.02 & 20.45 & 18.82 & 15.70 & 17.39 \\
23 & InternVL3.5-8B & 15.56 & 15.04 & 20.13 & 15.29 & 16.70 & 16.51 \\
24 & GLM-4.6V & 15.56 & 8.69 & 19.81 & 21.18 & 14.10 & 16.31 \\
25 & LLaVA-Video-72B & 13.33 & 15.04 & 18.83 & 17.65 & 16.20 & 16.21 \\
26 & Qwen3-VL-4B & 14.81 & 8.69 & 19.48 & 21.18 & 13.90 & 16.04 \\
27 & Keye-VL-1.5-8B & 14.81 & 12.92 & 20.45 & 14.12 & 15.60 & 15.58 \\
28 & InternVL3.5-30B-A3B & 12.59 & 16.10 & 21.75 & 11.76 & 17.00 & 15.55 \\
29 & Qwen3-VL-8B & 15.56 & 10.59 & 17.86 & 16.47 & 14.00 & 15.12 \\
30 & InternVL3.5-4B & 14.81 & 13.14 & 17.86 & 7.06 & 14.30 & 13.22 \\
31 & GLM-4.6V-Flash & 11.11 & 7.20 & 14.29 & 20.00 & 11.00 & 13.15 \\
32 & VITA-1.5 & 19.26 & 8.49 & 13.42 & 10.59 & 10.60 & 12.94 \\
33 & LLaVA-Video-7B & 11.11 & 13.14 & 13.96 & 7.06 & 12.60 & 11.32 \\
34 & VideoLLaMA3-7B & 7.52 & 13.69 & 11.72 & 8.33 & 11.80 & 10.32 \\
\bottomrule
\end{tabular}
\end{table}

%% file: tables/subclass_result.tex
\begin{table*}[t]
\centering
\small
\setlength{\tabcolsep}{5pt}
\renewcommand{\arraystretch}{1.05}
\caption{
Average subcategory-level accuracy of all models on \ourmethod.
}
\label{tab:subclass_average_accuracy}
\resizebox{\textwidth}{!}{
\begin{tabular}{llcc|llcc}
\toprule
\textbf{Domain} & \textbf{Subclass} & \textbf{\#Samples} & \textbf{Avg. Acc.(\%)}
& \textbf{Domain} & \textbf{Subclass} & \textbf{\#Samples} & \textbf{Avg. Acc.(\%)} \\
\midrule
AIGC & Artifacts & 79 & 11.91
& Games & Music & 20 & 10.00 \\

GUI & XR & 8 & 26.52
& Human & Basketball & 57 & 16.84 \\

GUI & Website & 97 & 17.32
& Human & Soccer & 38 & 13.51 \\

GUI & Application & 91 & 15.11
& Human & Volleyball & 40 & 12.57 \\

GUI & Operating System & 22 & 31.29
& Human & Other Ball Sports & 37 & 14.00 \\

Nature & Lightning & 49 & 45.52
& Human & Non-ball Sports & 40 & 24.78 \\

Nature & Other Natural Phenomena & 23 & 30.95
& Human & Daily Life & 38 & 8.59 \\

Industry & Manual Assembly & 96 & 16.33
& Animal & Birds & 42 & 35.08 \\

Industry & Industrial Machinery & 25 & 13.53
& Animal & Mammals & 13 & 32.58 \\

Games & FPS & 60 & 13.70
& Animal & Reptiles & 29 & 34.69 \\

Games & MOBA & 50 & 17.06
& Animal & Amphibians & 6 & 24.51 \\

Games & Combat & 30 & 22.84
& Animal & Fish & 2 & 73.53 \\

Animal & Invertebrates & 8 & 51.84
& & & \\
\bottomrule
\end{tabular}
}
\end{table*}

%% file: figures/heatmap_subclass.tex
\begin{figure*}[t]
    \centering
    \includegraphics[width=0.88\textwidth]{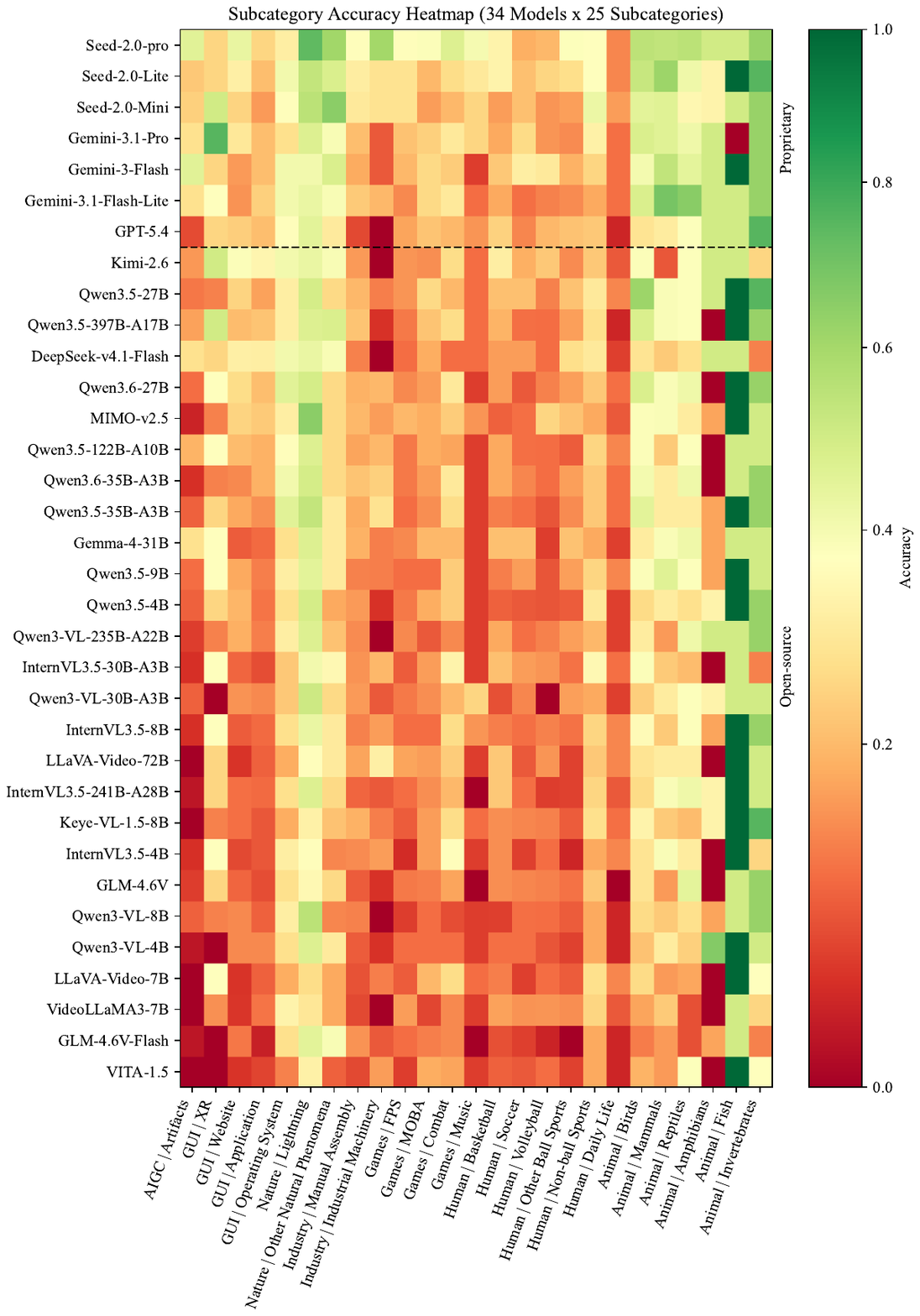}
    \caption{
    Subcategory-level accuracy heatmap on \textbf{\ourmethod}. Rows correspond to 34 evaluated models and columns correspond to 25 fine-grained subcategories. The color indicates accuracy, and the dashed horizontal line separates proprietary and open-source models.
    }
    \label{fig:subcategory_heatmap}
\end{figure*}

%% file: figures/duration_subplot.tex
\begin{figure*}[t]
    \centering
    \includegraphics[width=0.98\textwidth]{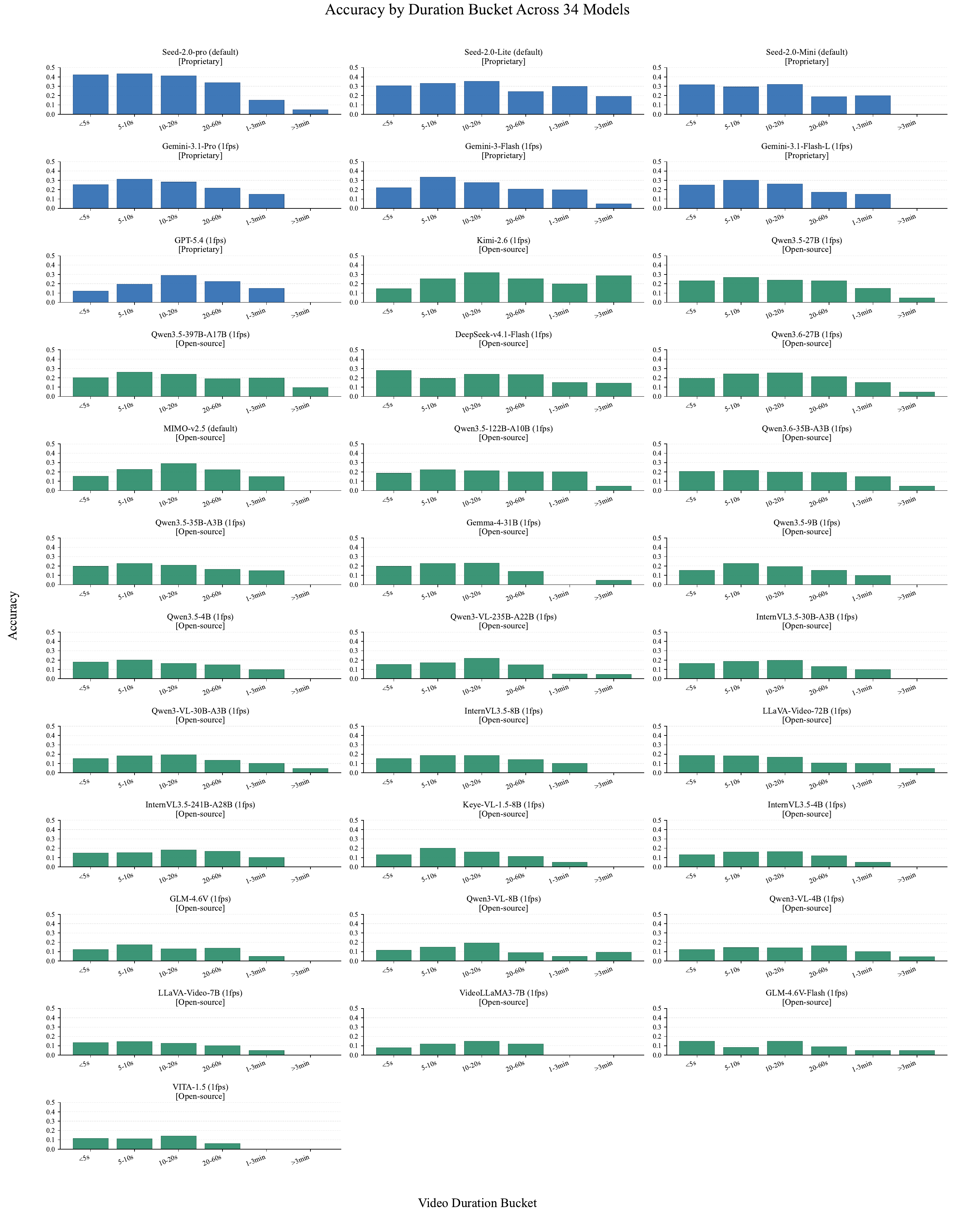}
    \caption{
    Duration-wise accuracy of all evaluated models on \textbf{\ourmethod}. Each subplot corresponds to one model, and bars represent accuracy across six video-duration buckets. The results show that most models perform better on short-to-medium videos and degrade when rapid events are embedded in longer temporal contexts.
    }
    \label{fig:duration_all_models}
\end{figure*}

%% file: tables/prompt_open.tex
\begin{figure}[t]
\centering
\begin{tcolorbox}[
    colback=gray!3,
    colframe=gray!45,
    title={Open-ended LLM-as-Judge Prompt},
    fonttitle=\bfseries,
    boxrule=0.5pt,
    arc=2pt,
    left=5pt,
    right=5pt,
    top=5pt,
    bottom=5pt,
    width=0.96\linewidth
]
\small
\textbf{System prompt:}

\medskip
\texttt{You are a strict and fair evaluator. Output valid JSON only.}

\medskip
\textbf{User message:}

\medskip
Evaluate whether the model answer is semantically consistent with the reference answer.

\medskip
\textbf{Judging rules:}
\begin{enumerate}
    \item Judge meaning, not wording.
    \item Accept paraphrases, synonyms, abbreviations, and equivalent naming variants.
    \item Mark as consistent if the model answer fully covers the reference answer's meaning, even with extra harmless details.
    \item Mark as inconsistent if it misses a key point, contradicts the reference answer, or changes important facts, including entity, action, order, quantity, identity, or existence.
    \item Mark as inconsistent if the model answer is too vague to support the same meaning.
    \item If uncertain, choose consistent only when a reasonable reader would conclude full semantic coverage.
\end{enumerate}

Return JSON only, with no markdown and no extra text:

\medskip
\texttt{\{"is\_consistent": true/false, "reason": "one-sentence explanation"\}}

\medskip
\textbf{Reference answer:}

\texttt{\{reference\_answer\}}

\medskip
\textbf{Model answer:}

\texttt{\{model\_answer\}}
\end{tcolorbox}
\caption{
Open-ended LLM-as-Judge prompt used to evaluate semantic consistency between the model prediction and the reference answer.
}
\label{fig:open_judge_prompt}
\end{figure}